\documentclass[lettersize,journal]{IEEEtran}
\usepackage[T1]{fontenc}
\usepackage{amsmath,amsfonts}
\usepackage{amssymb}
\usepackage{booktabs,array,xcolor,colortbl}
\definecolor{RetargetPrimary}{gray}{0.93}
\providecommand{\RetargetStat}[2]{%
  \makebox[2.7em][r]{$#1$}\,\ensuremath{\pm}\,\makebox[2.3em][r]{$#2$}}

\usepackage{algorithmic}
\usepackage{algorithm}
\usepackage{array}
\usepackage[caption=false,font=normalsize,labelfont=sf,textfont=sf]{subfig}
\usepackage{textcomp}
\usepackage{stfloats}
\usepackage{url}
\usepackage{verbatim}
\usepackage{graphicx}
\graphicspath{{kinematic_experiment/}}
\usepackage{cite}
\usepackage{booktabs}
\usepackage{multirow}
\usepackage{xcolor}
\usepackage{makecell}
\usepackage{textcase}
\usepackage{hyperref}
\newcommand{\cmark}{\textcolor{green!60!black}{\checkmark}}
\newcommand{\xmark}{\textcolor{red!75!black}{\times}}

\newcommand{\algname}{Morphometric Imitation}
\newcommand{\kinname}{Morphometric Optimization}
\newcommand{\algabbr}{MMO}
\newcommand{\arxivversion}{}

\title{\algname{}: From Morphology and Contact Aware Hand Retargeting to Sim-to-Real Visuomotor Policy}
\author{Tara~Sadjadpour, Siming~He, C.K.~Wolfe, Haozhi~Qi, Lea~Wilken, \\
S.~Shankar~Sastry, Claire~Tomlin$^{*}$, and~Jitendra~Malik$^{*}$%
\thanks{All authors are with the Department of Electrical Engineering and Computer Science at the University of California, Berkeley, CA, USA.}%
\thanks{$^{*}$Equal advising.}}

\begin{document}

\makeatletter
\g@addto@macro\@maketitle{%

    \setcounter{figure}{0}%
    \refstepcounter{figure}%
    \noindent\begin{minipage}{\textwidth}
        {\centering
        \includegraphics[width=1\textwidth]{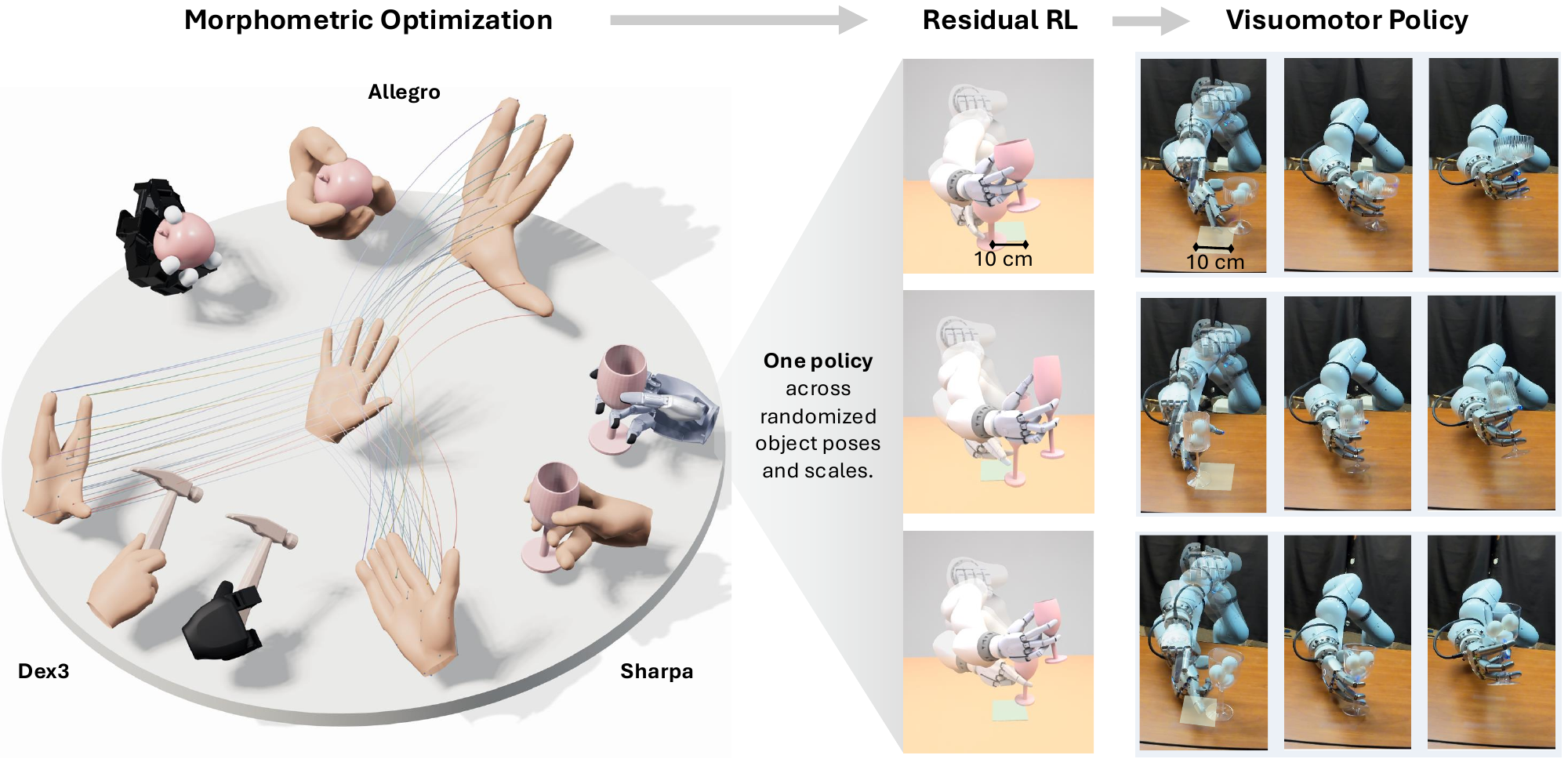}
        \par}

        {\footnotesize Fig.~\thefigure. \textbf{\algname{}.} 
        We present a three-stage framework that first kinematically retargets human hand-object trajectories to three-, four-, or five-fingered robot hands. Residual RL then dynamically retargets the kinematic reference into feasible demonstrations, which train a zero-shot sim-to-real visuomotor policy robust to object variation and initial poses. We illustrate the latter two stages with the Sharpa performing the wineglass demonstration. Residual RL rollouts are shown as time lapses across randomized object poses and scales. Visuomotor rollouts are shown across diverse wineglass instances, with the reach from home to pre-grasp shown as a time lapse.
        }
        \label{fig:teaser}
    \end{minipage}

    \vspace{-20pt}
}
\makeatother

\maketitle

\begin{abstract}
Human hand-object interactions (HOIs) provide a rich source of demonstrations for dexterous manipulation, but learning directly from them presents challenges in bridging morphology gaps, ensuring dynamical feasibility, and sim-to-real deployment.
We present \textit{\algname{}}, a three-stage framework that transforms reconstructed HOIs into zero-shot sim-to-real visuomotor policies. First, \MakeTextLowercase{\kinname{}} (MMO) kinematically retargets human motion across hand morphologies while preserving demonstrated contacts. Second, residual reinforcement learning (RL) refines the kinematic reference using object pose and contact information from the human motion to produce dynamically feasible robot demonstrations. Third, these demonstrations are distilled into visuomotor policies. 
Across three robot hands and ten HOIs, MMO outperforms five baselines in contact F1, improving on the strongest ones by 8 to 28 points, while improving the success rate of downstream dynamic retargeting by as much as 35 points.
On a Sharpa hand, the visuomotor policies achieve 89.3\% zero-shot success in 300 real-world trials on 30 objects spanning 10 categories.
\end{abstract}

\section{Introduction}\label{sec:intro}

\IEEEPARstart{H}{uman} motion data has gained increasing popularity as a data source for robot manipulation. 
While teleoperation provides high-quality robot demonstrations, it can be limited by cost, time to collect, and the availability of expert data collectors.
Consequently, recent works have explored human motion data to reduce the amount of teleoperated data required for robot learning. Some approaches use a combination of 2D human motion from monocular video for pretraining, followed by post-training with teleoperated robot demonstrations~\cite{zheng2026egoscale, dyna2026dyna2}. Others apply 3D reconstruction to monocular video and use the resulting human hand-object trajectories as the source of demonstration data. The typical pipeline for using 3D human motion data consists of three stages: kinematic retargeting of the reconstructed human motion to the target robot~\cite{handa2020dexpilot, qin2023anyteleop, kim2025pyroki, Yang2026OmniRetarget, wu2026toporetarget, feng2026minimalist}, dynamic retargeting into physically feasible robot demonstrations~\cite{pan2025spider, chen2024object, singh2024hand, lum2025crossing, mandi2025dexmachina, li2025maniptrans, Yang2026OmniRetarget, wu2026toporetarget, feng2026minimalist, sharma2026one}, and distillation of these demonstrations into a visuomotor policy~\cite{chen2024object, lum2025crossing, sharma2026one}.

\begin{figure*}[t]
    \centering
    \includegraphics[width=1\textwidth]{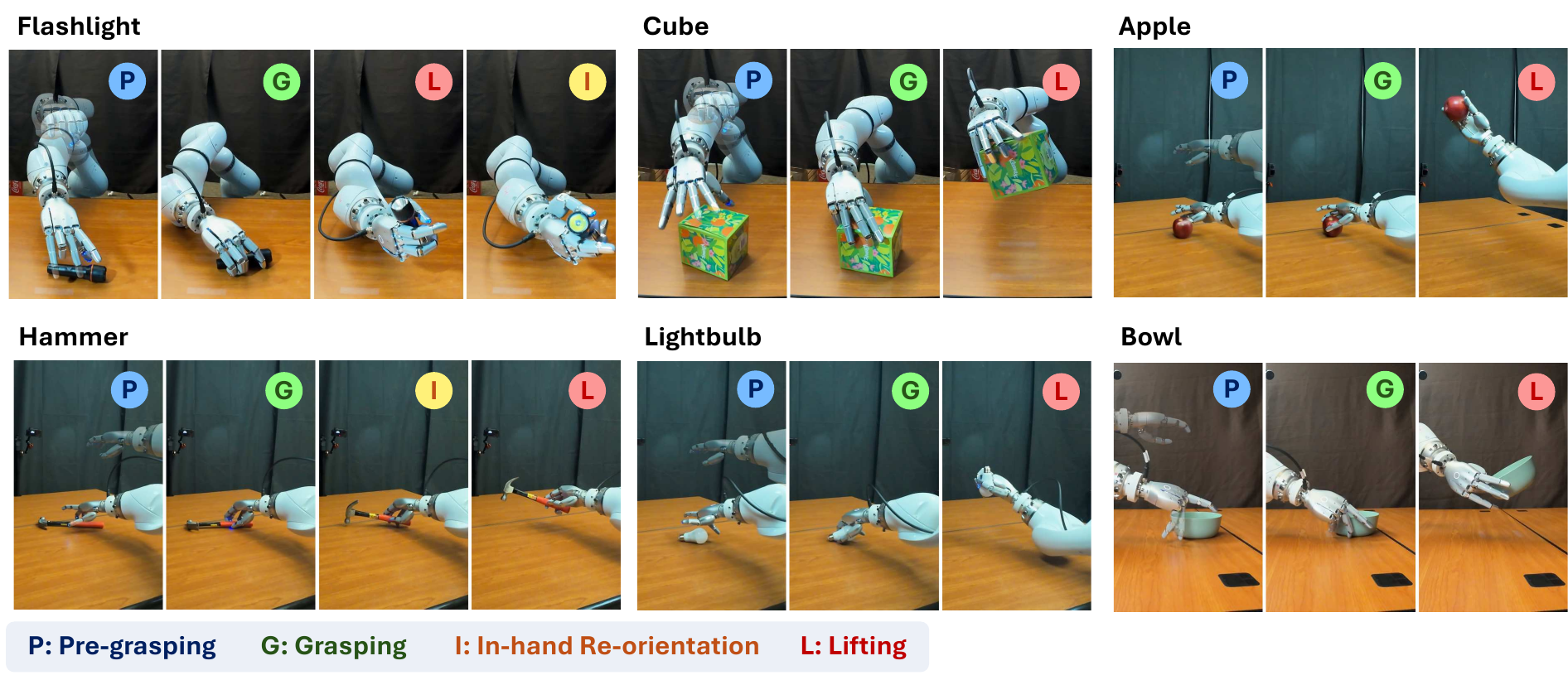}
    \caption{\textbf{Visuomotor Policy Rollouts.} Zero-shot sim-to-real rollouts for six objects: flashlight, cube, apple, hammer, lightbulb, and bowl. For each rollout, the first frame shows a time lapse of the reach from the home position to the pre-grasp, followed by the grasp in the second frame. The cube, apple, lightbulb, and bowl trajectories conclude with a lift. The flashlight trajectory additionally performs in-hand reorientation to point the flashlight forward, while the hammer trajectory reorients the hammer head toward the table before lifting.}
    \label{fig:real_experiments}
\end{figure*}

In this work, we introduce \textit{\algname{}}, a three-stage framework that transforms reconstructed human hand-object interactions into visuomotor imitation learning policies. The first stage is \MakeTextLowercase{\kinname{}} (\algabbr{}), a novel kinematic retargeting method that accounts for differences in human and robot hand morphology while preserving the demonstrated hand-object contacts. We then perform dynamic retargeting with residual reinforcement learning (RL) in simulation, jointly leveraging object pose and contact information from the reference motion. The residual policy adapts the kinematic reference to produce dynamically feasible trajectories with stable contact dynamics and reliable collision avoidance. Finally, we use these trajectories as demonstrations to train a visuomotor policy via imitation learning in simulation for zero-shot sim-to-real deployment.

Realizing the benefits of transferring human hand-object interactions to robotic hands presents unique challenges at each stage. During kinematic retargeting, differences in human and robot hand morphology and size make perfect geometric correspondence generally unattainable, resulting in discrepancies in hand-object contact locations. Then, dynamic retargeting must transform \textit{natural} human motion into physically feasible and safe robot motion. Some prior work facilitates this transfer by constraining human demonstrations to robot workspaces and favorable grasping strategies, or by using slow, controlled motions that simplify contact dynamics~\cite{lum2025crossing, kedia2026simtoolreal}. While effective for transfer, such constraints rely on purposefully collected human demonstrations and may not extend to the challenges present in naturally occurring human motion at scale. Finally, deploying visuomotor policies trained on dynamically retargeted demonstrations introduces the sim-to-real gap. Some approaches alleviate this challenge by learning from human motion in simulation and subsequently fine-tuning with real teleoperated robot data~\cite{singh2024hand}.

Our key insight is that morphology- and contact-aware retargeting improves both kinematic and dynamic retargeting, yielding physically feasible and safe robot demonstrations for training sim-to-real visuomotor policies. As summarized in Fig.~\ref{fig:teaser}, our framework proceeds from morphometric optimization to residual RL and, ultimately, zero-shot real-world visuomotor control. For kinematic retargeting, \algabbr{} aligns the human and robot hand morphologies, recovers the demonstrated hand-object contacts, and transfers the resulting morphology- and contact-aligned motion to the robot. For dynamic retargeting, we introduce a residual RL formulation that uses object pose and contact information in the observations, rewards, and termination conditions to preserve the demonstrated object motion and the contact behavior that produces it. Across both stages, we explicitly address robot-table collisions, which are particularly important when transferring natural tabletop human motions. The resulting dynamically feasible demonstrations are then distilled into visuomotor policies for zero-shot sim-to-real deployment.

We evaluate kinematic and dynamic retargeting on three robot hands and ten GRAB hand-object trajectories~\cite{taheri2020grab}. For kinematic retargeting, \algabbr{} outperforms five baselines~\cite{qin2023anyteleop, handa2020dexpilot, kim2025pyroki, Yang2026OmniRetarget} in contact preservation, improving F1 by at least 8 points over the strongest baseline with consistently lower mean patch distance (Table~\ref{tab:retarget-main}). This translates to dynamic retargeting, yielding higher task success on all three hands by as much as 35 points over the strongest baseline, more accurate object trajectory tracking, and final grasps closer to the demonstrated contacts on average across the three hands (Table~\ref{tab:retargeting_comparison}). 
Ablations show that object pose and contact information play complementary roles in dynamic retargeting.
Finally, our visuomotor policies achieve 89.3\% zero-shot success in
300 real-world trials on 30 objects from ten categories across diverse
initial poses (Fig.~\ref{fig:real_experiments}).

\begin{table*}[t]
\centering
\caption{Comparison of methods learning from human motion data. The first three columns indicate components proposed by each method. Natural human motion, marked only for dynamic retargeting methods, denotes learning solely from human demonstrations collected without constraints that facilitate robot control.}
\label{tab:method_comparison}
\footnotesize
\setlength{\tabcolsep}{6pt}
\begin{tabular}{lcccc}
\toprule
\textbf{Method}
& \textbf{Kinematic Retargeting}
& \textbf{Dynamic Retargeting}
& \textbf{\makecell{Zero-Shot Sim-to-Real\\Visuomotor Policy}}
& \textbf{\makecell{Natural Human\\Motion Data}}\\
\midrule
DexPilot~\cite{handa2020dexpilot} & $\cmark$ & $\xmark$ & $\xmark$ & $--$ \\
AnyTeleop~\cite{qin2023anyteleop} & $\cmark$ & $\xmark$ & $\xmark$ & $--$ \\
\texttt{Position}~\cite{qin2023anyteleop} & $\cmark$ & $\xmark$ & $\xmark$ & $--$ \\
Contact-Aware PyRoki~\cite{kim2025pyroki} & $\cmark$ & $\xmark$ & $\xmark$ & $--$ \\
SPIDER~\cite{pan2025spider} & $\xmark$ & $\cmark$ & $\xmark$ & $\cmark$ \\
Chen et al.~\cite{chen2024object} & $\xmark$ & $\cmark$ & $\cmark$ & $\cmark$ \\
HOP~\cite{singh2024hand} & $\xmark$ & $\cmark$ & $\xmark$ & $\xmark$ \\
Human2Sim2Robot~\cite{lum2025crossing} & $\xmark$ & $\cmark$ & $\cmark$ & $\xmark$ \\
DexMachina~\cite{mandi2025dexmachina} & $\xmark$ & $\cmark$ & $\xmark$ & $\cmark$ \\
ManipTrans~\cite{li2025maniptrans} & $\xmark$ & $\cmark$ & $\xmark$ & $\cmark$ \\
OmniRetarget~\cite{Yang2026OmniRetarget} & $\cmark$ & $\cmark$ & $\xmark$ & $\cmark$ \\
TopoRetarget~\cite{wu2026toporetarget} & $\cmark$ & $\cmark$ & $\xmark$ & $\cmark$ \\
REGRIND~\cite{feng2026minimalist} & $\cmark$ & $\cmark$ & $\xmark$ & $\cmark$ \\
DemoMimic~\cite{sharma2026one} & $\xmark$ & $\cmark$ & $\cmark$ & $\cmark$ \\
\textbf{\algname{} (Ours)} & $\cmark$ & $\cmark$ & $\cmark$ & $\cmark$ \\
\bottomrule
\end{tabular}
\end{table*}

\section{Related Work}\label{sec:related_work}
\subsection{Kinematic Retargeting}
In this work, we focus on unsupervised kinematic retargeting methods that do not require wearables.
Within this setting, prior approaches commonly formulate retargeting through hand keypoint vector constraints~\cite{qin2023anyteleop, sivakumar2022robotic, naughton2024respilot, handa2020dexpilot, kim2025pyroki}. 
Among these methods, the repository \texttt{Dex-Retargeting}~\cite{qin2023anyteleop} is widely used. Its vector-based formulation, introduced in AnyTeleop~\cite{qin2023anyteleop}, aligns relative robot fingertip positions with human fingertip vectors uniformly scaled by the robot-to-human hand size ratio; the accompanying codebase also provides a position-based formulation for offline retargeting, referred to as \texttt{Position}. DexPilot~\cite{handa2020dexpilot} predates AnyTeleop and similarly uses a vector-based formulation, with additional inter-finger constraints.
More recent methods incorporate object contact and morphology into kinematic retargeting. PyRoki~\cite{kim2025pyroki} extends the AnyTeleop formulation with a contact-aware objective based on fingertip-object proximity and optimizes link scales to account for hand morphology; it has since been adopted by works, such as~\cite{allshire2025visual}. OmniRetarget~\cite{Yang2026OmniRetarget} introduces contact-aware kinematic retargeting for locomanipulation through tetrahedral mesh matching and is widely adopted in that community. In contrast, our method optimizes the human hand model to match the morphology of the target robot and then recovers the hand-object contacts altered by this morphological transformation.


It is widely recognized that the quality of kinematic references impacts downstream dynamic retargeting performance~\cite{liao2026beyondmimic, zhang2025hub, Yang2026OmniRetarget}.
OmniRetarget~\cite{Yang2026OmniRetarget} evaluates retargeting with kinematic metrics
and with the performance of the same RL formulation trained on each method's trajectories. We adopt both evaluation protocols, as kinematic retargeting ultimately produces the references that residual RL refines into demonstrations for zero-shot sim-to-real visuomotor policies.

\textbf{Concurrent Work.} REGRIND~\cite{feng2026minimalist} and TopoRetarget~\cite{wu2026toporetarget} build on OmniRetarget~\cite{Yang2026OmniRetarget}, also incorporating an interaction mesh into their objectives. REGRIND does not demonstrate improvements over OmniRetarget. TopoRetarget reports improvements but had not released its code at the time of writing.

\subsection{Dynamic Retargeting}
Recent work has explored dynamic retargeting of hand-object interactions using physics simulation with RL~\cite{chen2024object, lum2025crossing, li2025maniptrans, mandi2025dexmachina, feng2026minimalist, sharma2026one}, optimization~\cite{pan2025spider}, or a combination of the two~\cite{singh2024hand}. 
These methods differ in their source demonstrations: some operate on natural human motion, while others incorporate teleoperated robot data~\cite{singh2024hand} or human demonstrations collected under constraints that facilitate robot transfer~\cite{lum2025crossing, kedia2026simtoolreal}. Our method solely uses unconstrained, \textit{natural} human motion; Table~\ref{tab:method_comparison} summarizes these and other distinctions.


These methods use the kinematic reference in different ways. Chen et al.~\cite{chen2024object} learn a residual policy over the retargeted wrist motion with only an object-tracking reward, then distill it into a zero-shot sim-to-real policy. Human2Sim2Robot~\cite{lum2025crossing} initializes the robot in a pre-grasp configuration and trains an RL policy with an object-tracking reward, and deploys a zero-shot sim-to-real visuomotor policy. SPIDER~\cite{pan2025spider} warm-starts sim-in-the-loop sampling-based model predictive control. ManipTrans~\cite{li2025maniptrans} tracks retargeted hand keypoints, DexMachina~\cite{mandi2025dexmachina} tracks both keypoints and joint angles from~\cite{qin2023anyteleop}, and HOP~\cite{singh2024hand} refines trajectories from~\cite{qin2023anyteleop} with sim-in-the-loop optimization, trains an RL policy to track hand keypoints, and fine-tunes it on teleoperated data.

\textbf{Concurrent Work.} REGRIND~\cite{feng2026minimalist} proposes an RL approach that encourages object tracking, while DemoMimic~\cite{sharma2026one} encourages both object tracking and contact alignment with the reference human motion. Unlike our focus on rigid object categories, DemoMimic targets articulated box trajectories for zero-shot sim-to-real deployment; its code was not available at the time of writing.
\section{Modeling Human Hands with MANO}\label{sec:bg}

Our method builds on the MANO hand model~\cite{romero2022embodied},
\begin{align}
    M(\overset{\rightarrow}{\beta}, \overset{\rightarrow}{\theta}) &= W(T_P(\overset{\rightarrow}{\beta}, \overset{\rightarrow}{\theta}),J(\overset{\rightarrow}{\beta}),\overset{\rightarrow}{\theta},\mathcal{W}),
\end{align}
where the skinning function $W$ poses a rest-pose mesh $T_P \in \mathbb{R}^{V \times 3}$ about the rest-pose joint locations $J \in \mathbb{R}^{(K+1) \times 3}$ according to the pose $\overset{\rightarrow}{\theta}$ and the per-vertex blend weights $\mathcal{W} \in \mathbb{R}^{V \times (K+1)}$. The rest-pose mesh and joints are obtained from a template hand mesh $\overline{\mathbf{T}} \in \mathbb{R}^{V \times 3}$ as
\begin{align}
    T_P(\overset{\rightarrow}{\beta}, \overset{\rightarrow}{\theta}) &= \overline{\mathbf{T}} + B_S(\overset{\rightarrow}{\beta}) + B_P(\overset{\rightarrow}{\theta}), \\
    J(\overset{\rightarrow}{\beta}) &= \mathcal{J} \big( \overline{\mathbf{T}} + B_S(\overset{\rightarrow}{\beta}) \big),
\end{align}
where $\mathcal{J} \in \mathbb{R}^{(K+1) \times V}$ is a joint regressor. The shape blend shapes $B_S(\overset{\rightarrow}{\beta}) = \sum_{n}\beta_n\mathbf{S}_n$ combine principal components $\mathbf{S}_n$ of hand shape with shape parameters $\overset{\rightarrow}{\beta}$, and the pose blend shapes
\begin{align}
    B_P(\overset{\rightarrow}{\theta}) &= \sum_{n=1}^{9K}\big(R_n(\overset{\rightarrow}{\theta})-R_n(\overset{\rightarrow}{\theta^*})\big)\mathbf{P}_n
\end{align}
weight corrective offsets $\mathbf{P}_n$ by the deviation of the joint rotations from the rest pose $\overset{\rightarrow}{\theta^*}$, avoiding the overly smooth deformations and joint collapse of standard linear blend skinning. The blend weights, blend shapes, and joint regressor are all learned from registered hand scans.

MANO has $K=15$ finger joints plus the wrist, the root of the kinematic chain. The pose $\overset{\rightarrow}{\theta} = [\overset{\rightarrow}{\theta}_g; \overset{\rightarrow}{\theta}_h]$ comprises the wrist's global orientation $\overset{\rightarrow}{\theta}_g \in \mathbb{R}^3$ in axis-angle form and local finger-joint rotations $\overset{\rightarrow}{\theta}_h \in \mathbb{R}^{3K}$. The skinning function outputs the posed mesh vertices $\mathbf{V} \in \mathbb{R}^{V \times 3}$ and joint locations $\mathbf{J}_{\text{posed}}$, which a translation $\overset{\rightarrow}{t} \in \mathbb{R}^3$ places in world coordinates.

\begin{figure*}
    \centering
    \includegraphics[width=1.0\textwidth]{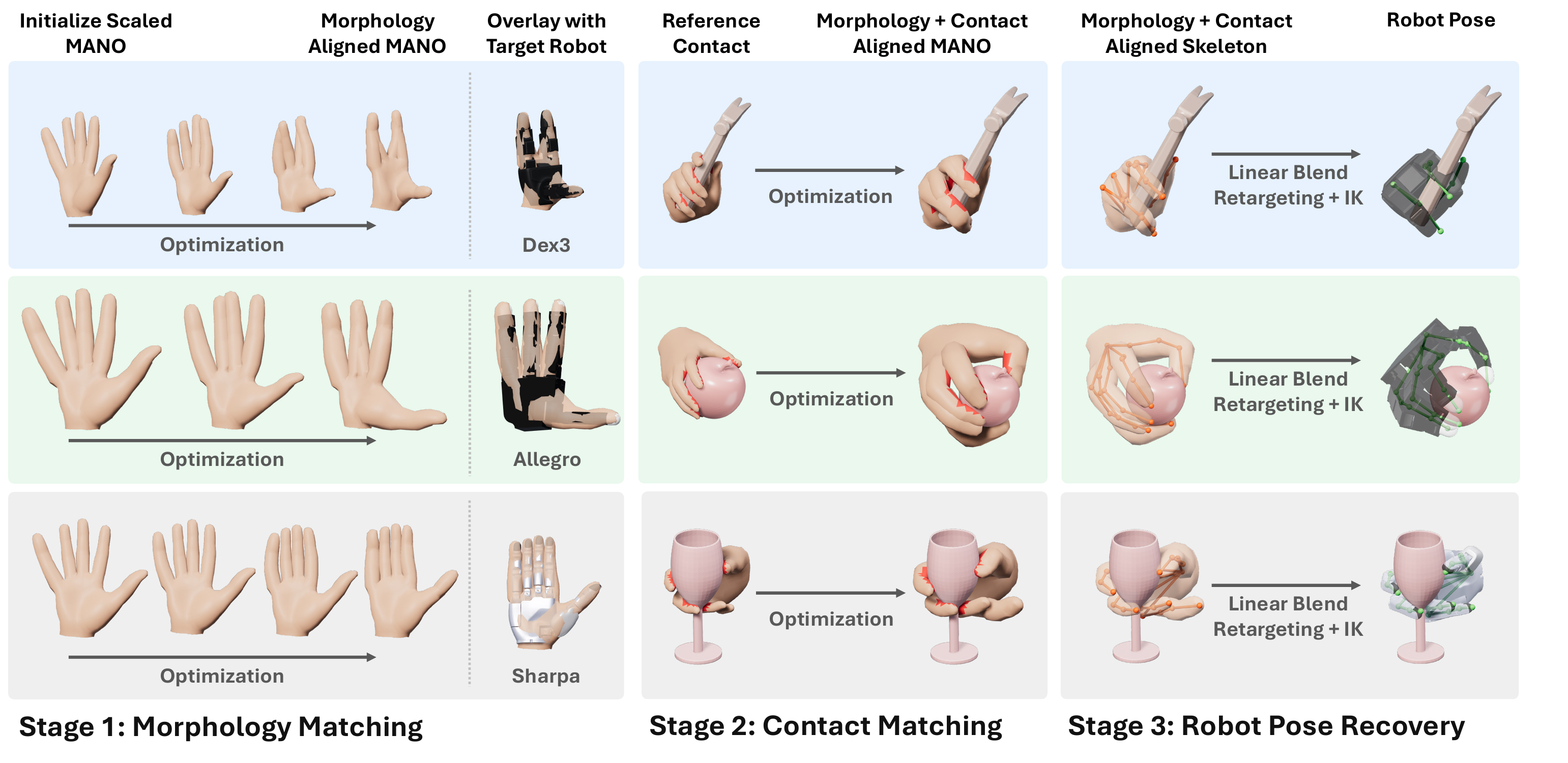}
    \caption{\textbf{\kinname{}.} We illustrate the three stages of morphometric optimization across three diverse robot hand embodiments: Dex3, Allegro, and Sharpa. (1) \textit{Morphology matching} initializes Scaled MANO using robot-to-human palm and finger-length ratios, then optimizes its morphology to match the target robot hand. (2) \textit{Contact matching} optimizes the morphology-aligned hand to recover the contacts from the reference human hand-object interaction. (3) \textit{Robot pose recovery} applies linear blend retargeting to transfer the resulting morphology- and contact-aligned MANO skeleton to the robot skeleton, followed by inverse kinematics to recover the robot joint configuration.}
    \label{fig:main_mmo}
\end{figure*}

\section{Kinematic Retargeting: \kinname{}}\label{sec:method}
The first stage of our pipeline converts human hand motion into robot hand motion with \textit{\kinname{} (\algabbr{})}, which consists of three steps (Figure~\ref{fig:main_mmo}). (1) \textit{Morphology matching} optimizes the MANO model to match the morphology of the robot hand. (2) Because this morphological change can alter the hand-object contacts, \textit{contact matching} re-optimizes the morphology-aligned MANO hand at each frame to recover the contacts of the original MANO hand. (3) \textit{Robot pose recovery} maps the resulting morphology- and contact-aligned MANO skeleton to the robot skeleton with \textit{linear blend retargeting}, which adapts the blending principle of linear blend skinning (LBS)~\cite{baran2007automatic}, and then solves inverse kinematics for the robot joint configuration. These steps are sequential dependencies rather than interchangeable modules: linear blend retargeting requires a MANO skeleton aligned to the robot skeleton, which morphology matching provides, and contact matching operates on the morphology-aligned hand.

\subsection{Morphology Matching}


\textbf{Scaled MANO model.} 
The MANO model $M$ is limited to natural variations in human hand shape and cannot capture the large, part-specific changes in proporitions required to retarget across diverse hand morphologies. Therefore, we extend it to the \textit{Scaled MANO model} $M_s$, with scaling vector $\overset{\rightarrow}{S}\in\mathbb{R}^6$ to independently scale the palm and each finger. This enables substantially greater variation in hand proportions.

From the robot's URDF, we set the palm scale $S_{\text{palm}}$ to the ratio between the robot's and the unscaled MANO hand's mean distance from the wrist to the root joint of each finger, known as its metacarpophalangeal (MCP) joint, and each finger scale $S_i$ to the corresponding ratio of MCP-to-fingertip distances. $M_s$ replaces the template mesh $\overline{\mathbf{T}}$, shape blend shapes, and pose blend shapes with scaled counterparts, while leaving the blend weights $\mathcal{W}$ unchanged. Because the blend weights of palm vertices are distributed across the wrist and multiple MCP joints, the palm region cannot be isolated cleanly, so we scale in two steps.

In Step 1, we scale the entire hand uniformly by $S_{\text{palm}}$ about the wrist joint position $\mathbf{j}_{\text{wrist}} = \mathcal{J}_{\text{wrist}} \cdot \overline{\mathbf{T}}$, where $\mathcal{J}_{\text{wrist}}$ is the wrist row of the joint regressor: $\overline{\mathbf{T}}' = \mathbf{j}_{\text{wrist}} + S_{\text{palm}} ( \overline{\mathbf{T}} - \mathbf{j}_{\text{wrist}} )$, $\mathbf{S}'_n = S_{\text{palm}} \, \mathbf{S}_n$, and $\mathbf{P}'_n = S_{\text{palm}} \, \mathbf{P}_n$ for all $n$.

In Step 2, we adjust each finger's length independently about its MCP joint. For finger $i \in \{\text{index}, \text{middle}, \text{pinky}, \text{ring}, \text{thumb}\}$, let $\mathbf{j}_i^{\text{MCP}} = \mathcal{J}_i^{\text{MCP}} \cdot \overline{\mathbf{T}}'$ be its MCP joint position regressed from the palm-scaled template, $\mathcal{K}_i$ the indices of its MCP, proximal interphalangeal (PIP), and distal interphalangeal (DIP) joints, and $\alpha_i = S_i / S_{\text{palm}}$ its scale relative to the palm. The finger membership weight $w_i(v) = \sum_{k \in \mathcal{K}_i} \mathcal{W}_{v,k}$ sums the blend weights of vertex $v$ over the joints of finger $i$, so $w_i(v) \approx 1$ for the vertices $v \in V_i$ of finger $i$ and $w_i(v) \approx 0$ otherwise. We update each template vertex $v$ as
\begin{align}
    \overline{\mathbf{T}}''_v &= \overline{\mathbf{T}}'_v + w_i(v) \left(\alpha_i - 1\right) \left(\overline{\mathbf{T}}'_v - \mathbf{j}_i^{\text{MCP}}\right),
\end{align}
which simplifies to $\overline{\mathbf{T}}''_v \approx \alpha_i (\overline{\mathbf{T}}'_v - \mathbf{j}_i^{\text{MCP}}) + \mathbf{j}_i^{\text{MCP}}$ for $v\in V_i$ and $\overline{\mathbf{T}}''_v \approx \overline{\mathbf{T}}'_v$ otherwise: each finger is scaled about its MCP joint without affecting the palm or other fingers. The shape and pose blend shapes are updated likewise, $\mathbf{S}''_{n,v} = \mathbf{S}'_{n,v} (1 + w_i(v) (\alpha_i - 1))$ and $\mathbf{P}''_{n,v} = \mathbf{P}'_{n,v} (1 + w_i(v) (\alpha_i - 1))$.

The resulting Scaled MANO model is
\begin{align}
    M_s(\overset{\rightarrow}{\beta}, \overset{\rightarrow}{\theta}, \overset{\rightarrow}{S}) &= W\!\left(T_P^s(\overset{\rightarrow}{\beta}, \overset{\rightarrow}{\theta}),\, J^s(\overset{\rightarrow}{\beta}),\, \overset{\rightarrow}{\theta},\, \mathcal{W}\right)
\end{align}
where $T_P^s$ uses $\overline{\mathbf{T}}''$, $\mathbf{S}''_n$, and $\mathbf{P}''_n$ in place of their unscaled counterparts. Although the joint regressor $\mathcal{J}$ is unchanged, the rest-pose joints $J^s$ are scaled because they are regressed from the scaled template and shape blend shapes.



\textbf{Optimization.} From the robot's kinematic skeleton, we extract joint positions $\mathbf{J}_R \in \mathbb{R}^{N_j \times 3}$ and fingertip positions $\mathbf{F}_R \in \mathbb{R}^{N_f \times 3}$. A correspondence mapping $\mathcal{C}$ pairs semantically equivalent joints of the MANO and robot skeletons (e.g., the index PIP joints). Robot joints without a MANO counterpart are excluded; when correspondence requires a joint the robot lacks (e.g., when it has fewer joints per finger than MANO), we assign it a phantom position interpolated from neighboring joints. Through $\mathcal{C}$, we obtain the corresponding joint positions $\hat{\mathbf{J}}_s \in \mathbb{R}^{N_j \times 3}$ of $M_s$ and its fingertip positions $\mathbf{F}_s \in \mathbb{R}^{N_f \times 3}$ at designated vertices of its posed mesh $\mathbf{V}_s$.

We optimize $\Phi = [\overset{\rightarrow}{S}, \overset{\rightarrow}{\theta}_g, \overset{\rightarrow}{\theta}_h, \overset{\rightarrow}{t}]$ with the shape parameters fixed to $\overset{\rightarrow}{\beta} = \mathbf{0}$. We initialize $\overset{\rightarrow}{S}$ with the URDF-derived ratios above; $\overset{\rightarrow}{\theta}_g$ with the rotation aligning the MANO and robot palm frames, each constructed from orthogonal axes derived from the wrist, thumb tip, and middle fingertip positions; $\overset{\rightarrow}{\theta}_h$ with the flat zero pose; and $\overset{\rightarrow}{t}$ with the robot-to-MANO wrist displacement after the initial scaling and orientation.

Morphology matching then aligns the corresponding joints and fingertips of the scaled MANO and robot hands by solving the nonlinear least-squares problem
\begin{align}
    \mathcal{L}_{\text{align}} = w_j \cdot \mathcal{L}_j + w_f \cdot \mathcal{L}_f,
\end{align}
with scalar weights $w_j$, $w_f$ and joint and fingertip losses $\mathcal{L}_j = \frac{1}{N_j} \sum_{i=1}^{N_j} \| \hat{\mathbf{J}}_{s,i} - \mathbf{J}_{R,i} \|_2^2$ and $\mathcal{L}_f = \frac{1}{N_f} \sum_{k=1}^{N_f} \| \mathbf{F}_{s,k} - \mathbf{F}_{R,k} \|_2^2$.


We minimize $\mathcal{L}_{\text{align}}$ with Levenberg--Marquardt to obtain $\Phi^*$. Morphology matching is performed \textbf{once} per robot hand, yielding the morphology-aligned MANO hand $M_{\text{morph}}$.

\subsection{Contact Matching}
Changing the hand morphology can alter the contacts of the original human hand-object trajectory, so we re-optimize the pose of $M_{\text{morph}}$ at each timestep to recover them.


\textbf{Contact Detection.} 
At each timestep, we are given the reference human hand mesh, the object mesh, and the table plane. The contact set $\mathcal{H}_c \subset \{1,\ldots,V\}$ contains the hand vertices whose distance to the nearest object vertex is below $\tau=4.5\,\text{mm}$, the threshold recommended for the GRAB dataset~\cite{taheri2020grab}. The position $\mathbf{p}_v$ of each $v \in \mathcal{H}_c$ on the reference hand is its target contact position. Because the reference MANO hand and $M_{\text{morph}}$ share the same mesh topology, these vertex correspondences transfer directly (Figure~\ref{fig:main_mmo}).

Non-contact frames, such as the reach and retreat phases, have $\mathcal{H}_c^t=\emptyset$ and thus no contact target. To define a target throughout the sequence, we assign each non-contact frame the contact indices of the most-contacted frame $t^\star=\arg\max_t|\mathcal{H}_c^t|$, $\mathcal{H}_c^t\leftarrow\mathcal{H}_c^{t^\star}$, with target positions $\mathbf{p}_v^t$ taken from the reference hand at the current frame $t$. The contact term then tracks the region of the reference hand that will form the grasp, preserving the demonstrated pre-grasp during approach and providing a continuous target into contact.

\textbf{Coupled Fingers.} When the robot hand has fewer fingers than the human hand, $\mathcal{C}$ maps multiple MANO fingers to a single robot finger. These \textit{coupled} fingers must maintain their relative configuration throughout the motion. For each coupled pair $(i, j)$, we compute target distances $\mathbf{d}_{ij}^* \in \mathbb{R}^4$ between the three joints and fingertip of fingers $i$ and $j$ on $M_{\text{morph}}$.

\textbf{Optimization.} At each timestep $t$, we optimize $\Psi = [\overset{\rightarrow}{\theta}_g, \overset{\rightarrow}{t}, \overset{\rightarrow}{\theta}_h]$ with $\overset{\rightarrow}{S}$ fixed to its morphology-matched value. Frames are processed sequentially: $\Psi_t$ is initialized with the previous solution $\Psi_{t-1}^*$, and the first frame with the global orientation, translation, and hand pose of $\Phi^*$.

Contact matching solves the nonlinear least-squares problem
\begin{align}
    \mathcal{L}_{\text{contact}} = w_c \cdot \mathcal{L}_c + w_d \cdot \mathcal{L}_d + w_{\text{table}} \cdot \mathcal{L}_{\text{table}} + w_p \cdot \mathcal{L}_p
\end{align}
with the following terms.

The contact loss $\mathcal{L}_c$ aligns the contact vertices $\hat{\mathbf{p}}_v(\Psi_t)$ of the re-posed $M_{\text{morph}}$ with their reference positions:
\begin{align}
    \mathcal{L}_c = \frac{1}{|\mathcal{H}_c|} \sum_{v \in \mathcal{H}_c} \| \hat{\mathbf{p}}_v(\Psi) - \mathbf{p}_v \|_2^2.
\end{align}

The coupled finger distance loss $\mathcal{L}_d$ preserves the relative configuration of coupled finger pairs:
\begin{align}
    \mathcal{L}_d = \frac{1}{|\mathcal{G}|} \sum_{(i,j) \in \mathcal{G}} \sum_{l=1}^{4} \left( \| \hat{\mathbf{q}}_{i,l}(\Psi) - \hat{\mathbf{q}}_{j,l}(\Psi) \|_2 - d_{ij,l}^* \right)^2
\end{align}
where $\mathcal{G}$ is the set of coupled finger pairs and $\hat{\mathbf{q}}_{i,l}(\Psi_t)$ is the $l$-th joint or fingertip position of finger $i$. This term is omitted for five-fingered robot hands.

The table penetration loss $\mathcal{L}_{\text{table}} = \frac{1}{V} \sum_{v=1}^{V} [\max(0,\, z_{\text{table}} - \hat{z}_v(\Psi))]^2$ penalizes hand vertices below the table surface, where $z_{\text{table}}$ is the table height and $\hat{z}_v(\Psi_t)$ is the $z$-coordinate of vertex $v$. 
This term is omitted when all reference fingertips lie below
the table, as may occur when a humanoid's arms rest naturally
at its sides rather than interacting with the table.

The pose regularization loss $\mathcal{L}_p = \frac{1}{K+1} \sum_{k=0}^{K} \| \Delta \overset{\rightarrow}{\omega}_k \|_2^2$ penalizes deviations from the reference human hand pose and global orientation, where $\| \Delta \overset{\rightarrow}{\omega}_k \|_2$ is the geodesic distance on $\mathrm{SO}(3)$ between the optimized and reference rotations $\hat{R}_k(\Psi), R_k^* \in \mathrm{SO}(3)$ of the $k$-th joint.
We minimize $\mathcal{L}_{\text{contact}}$ with the Levenberg--Marquardt algorithm; the solution $\Psi_t^*$ gives the morphology- and contact-aligned MANO hand $M_{mc}^t$.



\subsection{Robot Pose Recovery}
Given $M_{mc}^t$, we recover the robot pose in two steps. Linear blend retargeting maps the aligned MANO kinematic skeleton to dense joint and fingertip targets on the robot skeleton, from which we construct 6-DoF pose targets and solve inverse kinematics (IK) for the robot joint configuration.



\textbf{Blend Weight Computation.} The blend weights are computed once, in the rest pose of $M_{\text{morph}}$. Let the robot's kinematic skeleton consist of $N$ joint and tip positions $\{\mathbf{r}_n\}_{n=1}^{N}$ connected by edges $\mathcal{E}$, and let $\{\mathbf{j}_k\}_{k=0}^{K}$ denote the $K{+}1$ joint and fingertip positions of $M_{\text{morph}}$. We compute blend weights $\mathbf{W} \in \mathbb{R}^{N \times (K+1)}$ with the heat diffusion method of~\cite{baran2007automatic}, adapted to a 1D skeleton graph by replacing its triangle-mesh cotangent Laplacian with the 1D cotangent Laplacian of~\cite{crane2019n}.



\textbf{Per-Frame Skinning.} 
At each timestep $t$, forward kinematics of $M_{mc}^t$ with pose parameters $\Psi_t^*$ gives the rigid transformation $G_k \in \mathrm{SE}(3)$ of each MANO joint $k$ and the relative transformation $\tilde{G}_k$ that maps points from the rest pose to the posed configuration. For each robot joint or fingertip $n$, we blend these transformations with the precomputed weights $W_{n,k}$ and apply the result to its rest-pose position $\mathbf{r}_n$:
\begin{align}
    \tilde{G}_n^{\text{blend}} &= \sum_{k=0}^{K} W_{n,k} \, \tilde{G}_k \\
    \hat{\mathbf{r}}_n &= \tilde{G}_n^{\text{blend}} \begin{pmatrix} \mathbf{r}_n \\ 1 \end{pmatrix} 
\end{align}
Linear blend retargeting thus yields Cartesian position targets for every robot joint and fingertip, including the fixed wrist joint that attaches the hand to a potential robot arm. With the parent-child connectivity of the robot kinematic tree, these positions directly define full 6-DoF pose targets for every hand link, including the wrist: the local $z$-axis points from each joint to its child, and Gram-Schmidt orthogonalization recovers the remaining axes. In contrast, prior retargeting methods typically recover only sparse fingertip position targets and require additional heuristics or optimization to infer even a wrist pose target for IK~\cite{shaw2024learning,qin2023anyteleop}; \algabbr{} needs no such separate pose reconstruction stage.

We use the IK solver in PyRoki~\cite{kim2025pyroki}, which optimizes the joint configuration $q$ via nonlinear least squares; in our arm-equipped experiments, $q$ contains both the arm and hand joints, which are optimized jointly.
Using capsule approximations and fixed-weight least-squares penalties for the collisions proved insufficient for hand-table interactions, so we add a JAX-optimized collision check that samples link surfaces while keeping IK fast. The IK objective includes costs for target position and orientation, self-collision, joint limits, joint velocities, and hand-table collisions, and is solved sequentially, initializing each timestep with the previous solution for temporal consistency.

\subsection{Implementation}
\algabbr{} is implemented in JAX using the MANO implementation of~\cite{yi2025estimating} and the open-source \texttt{jaxls} least-squares solver. On an NVIDIA RTX 4090 GPU, it retargets 180 frames per second, including the preprocessing that computes hand-object contact indices and excluding the IK step, which PyRoki's solver completes efficiently in JAX~\cite{kim2025pyroki}. We use the \textbf{same hyperparameters} across all hands, except for the correspondence mapping $\mathcal{C}$, which depends on the number of fingers of the target robot hand. Additional details are provided in the Supplemental Material.

\section{Dynamic Retargeting: Residual Reinforcement Learning}\label{sec:residual_method}
Kinematic retargeting provides a reference robot trajectory $\mathbf{q}_{\text{ref}}$ that captures the desired hand motion and contact behavior but does not account for the dynamics of the robot-object interaction. Therefore, we perform dynamic retargeting with residual RL in the ManiSkill simulator~\cite{taomaniskill3}. Using $\mathbf{q}_{\text{ref}}$ as a warm start, a policy $\pi$ trained with PPO predicts a residual action $\Delta\mathbf{q}_t$ at each timestep $t$, and the commanded joint configuration is $\mathbf{q}_{\text{cmd}} = \mathbf{q}_{\text{ref}} + \Delta\mathbf{q}$.
The policy thus adapts the kinematic reference to stable contact dynamics while remaining grounded in it, and its trajectories serve as ``ground-truth'' demonstrations for downstream policy distillation.

As a teacher policy, $\pi$ has access to privileged state information during training. Our central premise is that successful dynamic retargeting should preserve two complementary aspects of the reference interaction: the motion of the manipulated object and the contact behavior that produces it. The observation space, reward function, and termination conditions therefore reflect both object pose and contact information. They also maintain the hand-table collision avoidance of contact matching and IK: the observations include robot-table clearance, and rollouts terminate upon table collision.

\textbf{Observation Space.} 
At timestep $t$, the observation $o_t$ includes the proprioception $\mathbf{q}_t$, observed object pose $T_{\text{obs},t}$, reference object pose $T_{\text{ref},t}$, current and future reference joint configurations $\{\mathbf{q}_{\text{ref},\tau}\}_{\tau=t}^{t+H}$, observed robot-object contact state $C_{\text{obs},t}$, future reference contact states $\{C_{\text{ref},\tau}\}_{\tau=t}^{t+H}$, minimum robot link and table heights $(z_{\min}, z_{\text{table}})$, object properties (scale relative to reference mesh for three axes, mass, friction coefficients, and center of mass), table friction coefficients, and the previous action and joint configuration $(a_{t-1}, \mathbf{q}_{t-1})$, where $a_{t-1}$ is $\mathbf{q}_{\text{cmd}}$ at timestep $t-1$. The contact state $C$ is a boolean indicating non-zero robot-object contact force for the observed state and hand-object proximity within the contact threshold for the reference state.

\textbf{Reward Function.}
The reward $r_t = r_{\text{obj},t} \cdot r_{\text{contact},t}$ jointly encourages object trajectory tracking and contact matching.

To track the reference object motion, $r_{\text{obj},t}$ uses the average point distance (ADD) pose metric. With points $\{p_k\}_{k=1}^{N_p}$ sampled from the object mesh, the ADD error between the reference and observed object poses is
\begin{align}\label{eq:add}
D_{N_p,t}
&= \frac{1}{N_p}\sum_{k=1}^{N_p}
\big|\big|
T_{\text{ref},t}\cdot p_k - T_{\text{obs},t}\cdot p_k
\big|\big|,
\end{align}
which captures both translational and rotational deviations. We define $r_{\text{obj},t} = e^{-\alpha D_{N_p,t}}$, where $\alpha$ controls the sensitivity of the reward to object pose error.

To preserve the reference contact behavior across robot hands with different numbers of fingers, we set the target number of contacting fingers to $N_{\text{goal},t} = \min(N_{\text{H},t}, N_{\max})$, where $N_{\text{H},t}$ is the number of contacting fingers in the human reference and $N_{\max}$ is the number of robot fingers, and define
\begin{align}
r_{\text{contact},t}
= 1 -
\frac{
\left|N_{\text{goal},t}-N_{\text{obs},t}\right|
}{N_{\max}},
\end{align}
where $N_{\text{obs},t}$ is the number of robot fingers with non-zero contact force on the object. Normalizing by $N_{\max}$ accounts for embodiment: a one-finger mismatch is a larger fraction of a three-fingered hand's contacts than of a five-fingered hand's. This term counts contacting fingers and does not constrain where on the object they make contact.


\textbf{Early Termination.}
Episodes terminate early on poor object tracking, contact mismatch, collision, or excessive contact force. For tracking and contact, we terminate when an exponential moving average (EMA) of the ADD error or of the contact mismatch exceeds its threshold, so that transient errors do not end an episode prematurely. Safety violations terminate immediately: a robot-table collision ($z_{\min} < z_{\text{table}}$) or a hand-object contact force above a threshold. These conditions discourage behaviors that could damage the robot, environment, or manipulated object during real-world execution.

\textbf{Domain Randomization.} We randomize the robot's PD gains and hand friction coefficients; the object's friction coefficients, center of mass, mass, and anisotropic scale; and, at initialization, the object's position within a $10\,\text{cm}\times10$\,cm region on the table, its yaw within $\pm15^\circ$, and the starting timestep of the reference trajectory. Randomizing the starting timestep exposes the policy to a wider range of hand-object configurations and improves robustness to variations in grasp and object pose. We also apply random external wrenches to the object to improve grasp robustness.

Center-of-mass randomization is non-trivial for objects with complex geometries, particularly under anisotropic scale randomization, which changes the object's geometry and support region. We therefore sample viable center-of-mass locations that preserve the stability of the object's initial resting configuration; the algorithm is given in the Supplemental Material.

\textbf{Training.} We train one residual RL policy per object category using the \textbf{same hyperparameters} across all hands and trajectories (see Supplemental Material for details). Training takes 60--90 minutes per policy on an NVIDIA RTX 4090.
\begin{figure}[h]
    \centering
    \includegraphics[width=1\linewidth]{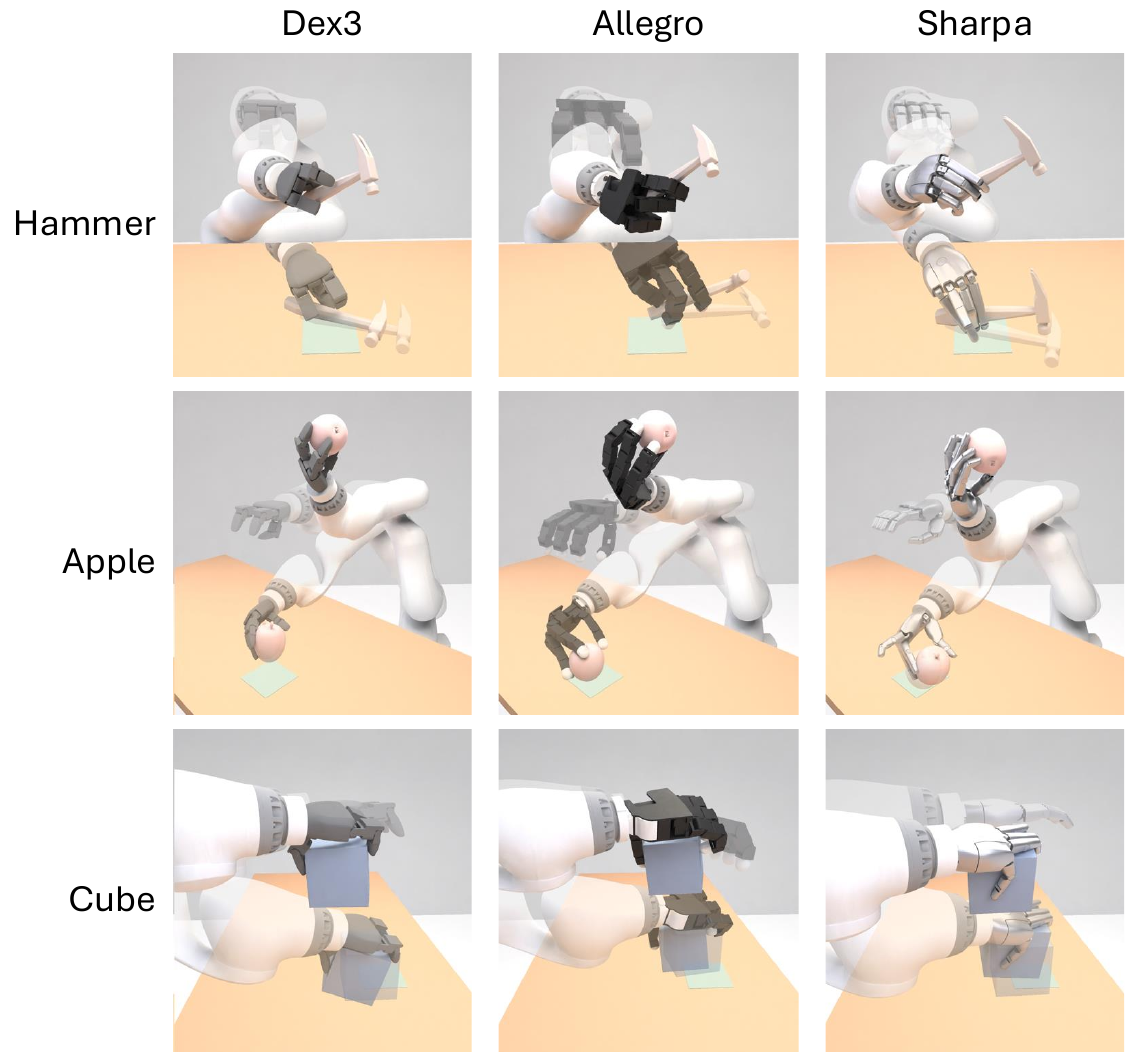}
    \caption{\textbf{Dynamic Retargeting Rollouts.} We show time lapses of residual RL policy rollouts for three robot hand embodiments (Dex3, Allegro, and Sharpa) across three object categories (hammer, apple, and cube). All policies use the same training hyperparameters, demonstrating that the residual RL formulation transfers across distinct hand morphologies and grasp configurations without embodiment- or task-specific tuning.}
    \label{fig:rl_rollouts}
\end{figure}

\section{Zero-Shot Sim-to-Real Visuomotor Policy}
\label{sec:distillation}
We distill each privileged residual RL teacher into a visuomotor student policy for zero-shot sim-to-real deployment. We collect demonstrations by rolling out the teacher under the domain randomizations above, except random timestep initialization, since demonstrations begin at the start of the reference trajectory. For each rollout, we record the commanded joint targets, the robot proprioception, and, at each timestep, a point cloud rendered from one simulated depth camera, from which RANSAC removes the table plane to retain only the robot and the manipulated object. The point clouds thus capture the variations in object geometry and pose induced by the scale and initialization randomization.


The student is supervised with the teacher's commanded joint targets and adopts ManiFlow's point-cloud encoder, DiT-X action generator, and joint flow-matching and consistency training~\cite{yan2025maniflow}. It is conditioned on the current and previous point clouds $(\mathbf{p}_t, \mathbf{p}_{t-1})$, current proprioception $\mathbf{q}_t$, and previous command $a_{t-1}$, and predicts chunks of joint targets, executing three actions at 10\,Hz before replanning.


During student training, we additionally randomize the simulated point clouds, complementing the physical variations in the teacher demonstrations and improving robustness to real-world depth observations.

\textbf{Training.} We train one visuomotor policy per object category using 10k demonstrations from its corresponding residual RL policy, with the \textbf{same hyperparameters} across all trajectories. Point cloud randomization and additional student policy training details are provided in the Supplemental Material.



\section{Experiments}\label{sec:experiments}
We evaluate \algname{} with four questions:
\begin{enumerate}
    \item Does \algabbr{} better preserve demonstrated hand-object contacts than prior kinematic retargeting methods across robot hand embodiments?
    \item Do the kinematic references from \algabbr{} lead to better downstream dynamic retargeting?
    \item Are both object pose and contact information important for successful dynamic retargeting?
    \item  How robustly do the resulting visuomotor policies transfer zero-shot to real-world objects with diverse geometries and physical properties across initial poses spanning the pose randomization region?
\end{enumerate}

\textbf{Evaluation Protocol.} 
We evaluate each stage of \algname{} through controlled comparisons appropriate to that stage. 
Following OmniRetarget~\cite{Yang2026OmniRetarget}, we evaluate kinematic retargeting both through contact fidelity directly and downstream dynamic retargeting performance by refining each method's reference with the same residual RL formulation, isolating the effect of the kinematic reference on task success.
For dynamic retargeting, we ablate object pose and contact information; the object pose-only variant captures the object-centric approaches commonly used in prior work~\cite{chen2024object, lum2025crossing, pan2025spider}, allowing us to isolate the contribution of contact information. Finally, we evaluate zero-shot sim-to-real transfer directly on hardware. We do not compare sim-to-real across methods, as differences in hardware, perception, and policy design would confound the comparison.

We organize the evaluation accordingly: after describing the experimental setup (Section~\ref{sec:eval-setup}), metrics (Section~\ref{sec:eval-metrics}), and baselines (Section~\ref{sec:baselines}), we present the results for each question in Sections~\ref{sec:kinematic_experiments}--\ref{sec:distillation_experiments}.


\subsection{Experimental Setup}\label{sec:eval-setup}
\textbf{Simulation Setup.} We use ten GRAB~\cite{taheri2020grab} hand-object trajectories (alarm clock, apple, bowl, large cube, cup, flashlight, hammer, lightbulb, torus, and wineglass) and three robot hands with three, four, and five fingers: Dex3, Allegro, and Sharpa. Each trajectory begins with the object resting on a table, followed by the human reaching, grasping, and lifting it. Some trajectories add task-specific object motion, such as bringing the cup toward the mouth, raising the wineglass in a toast, reorienting the flashlight to point forward, or rotating the hammer so that its head faces downward (see Fig.~\ref{fig:real_experiments}).

\textbf{Real-World Setup.} We deploy one visuomotor policy per object category on a KUKA iiwa14 arm with a Sharpa Wave hand and an Intel RealSense L515 LiDAR camera. The robot operates directly over a hardwood table without a compliant surface, making table collisions a highly consequential failure mode. Each category is tested on three physical instances (Figure~\ref{fig:real_world_objects}) that vary in shape, size, mass, center of mass, friction, and LiDAR observability. The transparent wineglass is invisible to the LiDAR camera, so we place ping pong balls in its bowl for partial observability (examples in the Supplemental Material). 
We evaluate each physical instance at 10 poses. Nine are fixed at the center, corners, and edge midpoints of the pose randomization region, emphasizing the challenging training pose distribution boundaries while enabling controlled analysis of spatial failure patterns. The tenth pose provides an additional stress test: for objects not rotationally symmetric about the yaw axis, we sample a random position with a yaw offset of at least $15^\circ$; for yaw-symmetric objects, we instead place the object slightly outside the pose randomization region. This yields 30 trials per category and 300 hardware trials total.


\subsection{Evaluation Metrics}
\label{sec:eval-metrics}
\textbf{Kinematic Metrics.} 
We measure how well the retargeted hand recovers the demonstrated contact geometry using location-aware F1 and contact patch distance. Our location-aware contact metrics require the robot to reproduce not only which hand part contacts the object, but also where on the object that contact occurs. Precision and recall for the F1 score are reported in the Supplemental Material.

For one trajectory, let $t$ index frames, $k$ index robot fingers and the palm, and $x_v$ denote the position of vertex $v$ on the object mesh, with all geometry expressed in object coordinates. Let $d^H_{tv}$ be the unsigned distance from $x_v$ to the MANO hand surface, $d^B_{tkv}$ its unsigned distance to the collision shapes of robot part $k$, $h^*_{tv}$ the human hand part closest to $x_v$, and $g$ the fixed mapping from human parts to robot parts. At contact tolerance $\tau$, the desired human and predicted robot contact indicators are $H_{tkv}=\mathbf{1}[d^H_{tv}\leq\tau] \cdot \mathbf{1}[g(h^*_{tv})=k]$ and $B_{tkv}=\mathbf{1}[d^B_{tkv}\leq\tau]$, so a correct robot contact must agree with the demonstration in frame, object location, and hand part. Weighting each vertex by its represented surface area $a_v$, since object meshes are nonuniformly tessellated, we accumulate contact area over all frames and parts:
\begin{equation}
\label{eq:retarget-contact-counts}
\begin{aligned}
\mathrm{TP}&=\sum_{t,k,v}a_vH_{tkv}B_{tkv},\\
\mathrm{FP}&=\sum_{t,k,v}a_v(1-H_{tkv})B_{tkv},\\
\mathrm{FN}&=\sum_{t,k,v}a_vH_{tkv}(1-B_{tkv}),
\end{aligned}
\end{equation}
from which precision, recall, and F1 score metrics follow using standard formulae. Precision penalizes robot contact outside the demonstrated patch, including contact in frames where the human makes none; recall penalizes demonstrated contact the robot fails to recover; and F1, the primary contact score, penalizes both. A method using hard constraints may fail to produce an output for a sequence, so averaging only over its successful outputs would inflate its scores. Thus, we report failure-adjusted scores: each contact metric is computed per sequence, set to zero for sequences the method fails to retarget, and averaged over all attempted sequences for each hand; patch distance, for which zero would be the best value, is instead averaged over the sequences that every method retargets successfully.

Contact patch distance measures spatial error continuously rather than thresholding the robot distance. Fixing the desired human patch at $\tau=5\,\mathrm{mm}$, $H^5_{tkv}=H_{tkv}(5\,\mathrm{mm})$, we compute
\begin{equation}
\label{eq:retarget-patch-distance}
D_{\rm patch}=\frac{\sum_{t,k,v}a_vH^5_{tkv}d^B_{tkv}}
                         {\sum_{t,k,v}a_vH^5_{tkv}},
\end{equation}
the area-weighted mean distance from every demonstrated contact location to the corresponding robot part, including locations the robot never reaches. Unlike recall, $D_{\rm patch}$ measures the magnitude of the error when a desired contact is missed, and fixing the human patch makes it independent of the robot contact tolerance. The main evaluation uses $\tau=5\,\mathrm{mm}$; the Supplemental also reports $\tau=1$ and $10\,\mathrm{mm}$. 

\textbf{Residual RL Metrics.}
We evaluate dynamic retargeting with contact-aware ADD (C-ADD) and task success rate (SR). C-ADD is Equation~\ref{eq:add} restricted to timesteps in which the robot or the reference human hand is in contact with the object, so that the stationary object during the reaching phase does not lower the tracking error. A rollout is successful if it (1) does not trigger early termination, including robot-table collision or excessive hand-object contact force, (2) has hand-object contact at the final timestep, and (3) achieves a final-frame ADD below $0.05$ relative to the reference object pose.
To measure whether the dynamically retargeted grasps preserve the demonstrated contacts, we additionally report contact patch distance at the final rollout state. We also report contact F1, precision, and recall in the Supplemental.

\textbf{Visuomotor Policy Metrics.} We report SR for the teacher and student in simulation and the student on hardware. A real-world trial is successful if (1) the robot does not undergo hard collisions with the table, (2) it lifts the object at least $5\,\mathrm{cm}$ above the table and completes the expected task trajectory (see Fig.~\ref{fig:real_experiments} for examples), and (3) it maintains a stable grasp for at least $5\,\mathrm{s}$.

\subsection{Baselines}
\label{sec:baselines}
We compare against five open-source and widely adopted kinematic retargeting baselines: DexPilot~\cite{handa2020dexpilot}, AnyTeleop~\cite{qin2023anyteleop}, \texttt{Position Dex-Retargeting}~\cite{qin2023anyteleop} (\texttt{Position}), Contact-Aware PyRoki~\cite{kim2025pyroki} (Contact PyRoki), and OmniRetarget~\cite{Yang2026OmniRetarget}.
Like \algabbr{}, OmniRetarget and \texttt{Position} jointly optimize the arm and hand. For kinematic evaluation, DexPilot, AnyTeleop, and Contact PyRoki are evaluated using their native vector-based, hand-only outputs. For downstream dynamic retargeting, which requires arm joints, we augment these three baselines with arm IK made with their own released solvers that places the wrist at the demonstrated wrist position at each frame, while preserving their original hand retargeting solvers. Four of the baselines use uniform scaling, while two of them match contacts. Therefore, using them as baselines measures the benefits of \algabbr{}'s formulation to be morphology and contact aware.

\subsection{Kinematic Retargeting Evaluation}
\label{sec:kinematic_experiments}
\textbf{Contact Preservation.}
Table~\ref{tab:retarget-main} compares \algabbr{} with the five baselines across ten demonstrations per hand. \algabbr{} achieves the highest location-aware F1 and lowest contact patch distance for all three hands. At $5\,\mathrm{mm}$, it exceeds the strongest baseline in F1 by $8.3$, $27.8$, and $10.5$ percentage points for Allegro, Dex3, and Sharpa, respectively. It also reduces mean patch distance to $9.1$, $7.5$, and $9.7\,\mathrm{mm}$, compared with $18.7$, $18.9$, and $10.7\,\mathrm{mm}$ for the respective strongest baselines. 
Together, the F1 and patch distance results show that MMO more faithfully preserves the demonstrated contact geometry across robot hand embodiments. To assess whether these improvements are consistent across demonstrations, we perform paired bootstrap analysis over the ten demonstrations for the five baselines and three hands ($5\times3=15$). At the $5\,\mathrm{mm}$ tolerance, the 95\% confidence intervals for the mean MMO--baseline F1 difference do not include zero in 13 of 15 comparisons. The two exceptions are Sharpa compared to the \texttt{Position} and Contact PyRoki baselines (Supplemental Material). Accordingly, while MMO achieves the highest mean F1 on every hand, these two comparisons are less conclusive.


\begin{table*}[ht!]
\centering
\small
\caption{Kinematic retargeting contact preservation at $5\,\mathrm{mm}$ tolerance (protocol in Sec.~\ref{sec:eval-metrics}). Mean $\pm$ standard deviation over ten demonstrations per hand. Bold and underline: best and second-best per hand and metric.}
\label{tab:retarget-main}
\begingroup
\setlength{\tabcolsep}{2.3pt}
\renewcommand{\arraystretch}{1.18}
\begin{tabular*}{\textwidth}{@{\extracolsep{\fill}}lcccccc@{}}
\toprule
 & \multicolumn{2}{c}{\textbf{Dex3}} & \multicolumn{2}{c}{\textbf{Allegro}} & \multicolumn{2}{c}{\textbf{Sharpa}} \\
\cmidrule(lr){2-3} \cmidrule(lr){4-5} \cmidrule(lr){6-7}
\textbf{Method} & Patch $\downarrow$ & F1 $\uparrow$ & Patch $\downarrow$ & F1 $\uparrow$ & Patch $\downarrow$ & F1 $\uparrow$ \\
 & (mm) & (\%) & (mm) & (\%) & (mm) & (\%) \\
\midrule
OmniRetarget~\cite{Yang2026OmniRetarget} & \RetargetStat{26.9}{10.9} & \RetargetStat{7.6}{11.3} & \RetargetStat{21.2}{13.1} & \RetargetStat{12.8}{16.0} & \RetargetStat{21.1}{15.5} & \RetargetStat{14.5}{18.1} \\
DexPilot~\cite{handa2020dexpilot} & \RetargetStat{42.6}{7.4} & \RetargetStat{2.5}{4.0} & \RetargetStat{\underline{18.7}}{4.9} & \RetargetStat{13.1}{6.8} & \RetargetStat{12.6}{3.7} & \RetargetStat{21.9}{9.8} \\
\texttt{Position}~\cite{qin2023anyteleop} & \RetargetStat{\underline{18.9}}{6.4} & \RetargetStat{\underline{10.2}}{7.6} & \RetargetStat{26.3}{39.4} & \RetargetStat{\underline{22.0}}{10.8} & \RetargetStat{25.4}{39.0} & \RetargetStat{23.9}{17.7} \\
AnyTeleop~\cite{qin2023anyteleop} & \RetargetStat{42.1}{7.2} & \RetargetStat{2.7}{3.9} & \RetargetStat{19.0}{5.1} & \RetargetStat{13.2}{7.1} & \RetargetStat{13.3}{3.7} & \RetargetStat{21.0}{9.8} \\
Contact PyRoki~\cite{kim2025pyroki} & \RetargetStat{26.6}{10.3} & \RetargetStat{9.6}{12.8} & \RetargetStat{21.5}{6.1} & \RetargetStat{15.4}{11.0} & \RetargetStat{\underline{10.7}}{2.7} & \RetargetStat{\underline{26.7}}{16.1} \\
\midrule
\algabbr{} (Ours) & \RetargetStat{\mathbf{7.5}}{1.4} & \RetargetStat{\mathbf{38.0}}{8.9} & \RetargetStat{\mathbf{9.1}}{2.0} & \RetargetStat{\mathbf{30.3}}{7.4} & \RetargetStat{\mathbf{9.7}}{6.2} & \RetargetStat{\mathbf{37.2}}{16.6} \\
\bottomrule
\end{tabular*}
\endgroup
\end{table*}

\begin{figure}[t]
\centering
\includegraphics[width=\linewidth]{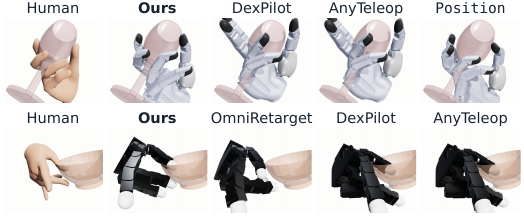}
\caption{\textbf{Kinematic Retargeting Outputs.} We show the final frames of the trajectories for wineglass with Sharpa (first row) and the bowl with Allegro (second row). Each row shows the human reference and the four methods with the highest F1 at the same frame. Our method more closely follows the demonstrated contact geometry than the baselines.
}
\label{fig:kinematic_comparison}
\end{figure}

\textbf{Qualitative Analysis.}
Figure~\ref{fig:kinematic_comparison} shows that, qualitatively, the robot hand retargeted by \algabbr{} more closely follows the demonstrated hand-object interaction than the baselines, including where on the object each finger makes contact. For both the wineglass and the bowl, \algabbr{} better preserves this contact geometry, which is central to reproducing the demonstrated interaction. At these frames, \algabbr{} achieves $38.5\%$ and $43.0\%$ F1, respectively, higher than all baselines.

\subsection{Dynamic Retargeting Evaluation}
\label{sec:dynamic_experiments}
We evaluate (1) how the choice of kinematic reference affects downstream dynamic retargeting and (2) the roles of object pose and contact information in our formulation. Each policy is evaluated over 2048 simulated rollouts per object category, with results averaged across the ten categories.
Figure~\ref{fig:rl_rollouts} shows rollouts on all three hands.


\textbf{Effect of Kinematic Reference.}
Table~\ref{tab:retargeting_comparison} compares the same residual RL formulation initialized from each kinematic retargeting method. \algabbr{} references yield the highest SR and lowest C-ADD across all three hands. SR reaches $82.5\%$, $69.9\%$, and $91.8\%$ for Dex3, Allegro, and Sharpa, compared with $53.2\%$, $68.2\%$, and $56.5\%$ for the strongest baseline on each hand. C-ADD similarly improves from $5.36$, $6.26$, and $5.76\,\mathrm{cm}$ to $4.10$, $5.01$, and $3.41\,\mathrm{cm}$. The residual RL formulation and hyperparameters are fixed across methods to isolate the effect of the kinematic reference. 
Paired bootstrap analysis across object categories shows that the 95\% confidence intervals for the mean \algabbr{}--baseline difference do not include zero in 11 of 15 SR and 10 of 15 C-ADD comparisons. The exceptions are concentrated in Allegro SR, where only Contact PyRoki does not include zero, and in C-ADD against OmniRetarget, where the intervals include zero for all three hands (see Supplemental Material).
Combined with the improved contact preservation in Table~\ref{tab:retarget-main}, these results indicate that the benefits of \algabbr{} extend to downstream physically feasible task execution.

Moreover, during evaluation, early termination occurs upon robot-table collision or excessive hand-object contact force. Averaged across the three hands, early termination rates are $5.2\%$ for \texttt{Position}, $6.9\%$ for \algabbr{}, $9.6\%$ for OmniRetarget, $11.8\%$ for AnyTeleop, $11.9\%$ for Contact PyRoki, and $12.5\%$ for DexPilot. These relatively small differences, together with the lower termination rate of \texttt{Position} than \algabbr{}, further suggest that the substantial gains in SR arise from differences in kinematic reference quality.
\begin{table*}[t]
\centering
\caption{Residual RL dynamic retargeting from different kinematic references. Mean $\pm$ standard deviation over ten categories, 2048 rollouts each (protocol in Sec.~\ref{sec:eval-metrics}). Bold and underline: best and second-best per hand and metric.}
\label{tab:retargeting_comparison}
\resizebox{\textwidth}{!}{%
\begingroup
\setlength{\tabcolsep}{3pt}
\renewcommand{\arraystretch}{1.18}
\begin{tabular}{lccccccccc}
\toprule
 & \multicolumn{3}{c}{\textbf{Dex3}} & \multicolumn{3}{c}{\textbf{Allegro}} & \multicolumn{3}{c}{\textbf{Sharpa}} \\
\cmidrule(lr){2-4} \cmidrule(lr){5-7} \cmidrule(lr){8-10}
\textbf{Method} & C-ADD $\downarrow$ & SR $\uparrow$ & Patch $\downarrow$ & C-ADD $\downarrow$ & SR $\uparrow$ & Patch $\downarrow$ & C-ADD $\downarrow$ & SR $\uparrow$ & Patch $\downarrow$ \\
 & ($10^{-2}$\,m) & (\%) & (mm) & ($10^{-2}$\,m) & (\%) & (mm) & ($10^{-2}$\,m) & (\%) & (mm) \\
\midrule
OmniRetarget~\cite{Yang2026OmniRetarget} & \RetargetStat{\underline{5.36}}{6.40} & \RetargetStat{\underline{53.2}}{38.9} & \RetargetStat{103.1}{137.8} & \RetargetStat{6.98}{7.11} & \RetargetStat{61.3}{41.8} & \RetargetStat{56.6}{50.7} & \RetargetStat{6.59}{6.25} & \RetargetStat{\underline{56.5}}{41.4} & \RetargetStat{\underline{75.5}}{101.7} \\
DexPilot~\cite{handa2020dexpilot} & \RetargetStat{10.7}{8.22} & \RetargetStat{17.3}{36.5} & \RetargetStat{126.0}{118.4} & \RetargetStat{9.12}{7.20} & \RetargetStat{39.1}{41.6} & \RetargetStat{108.2}{125.1} & \RetargetStat{10.7}{7.85} & \RetargetStat{27.3}{40.2} & \RetargetStat{118.7}{115.8} \\
\texttt{Position}~\cite{qin2023anyteleop} & \RetargetStat{7.32}{4.96} & \RetargetStat{39.8}{44.9} & \RetargetStat{\underline{58.4}}{60.7} & \RetargetStat{\underline{6.26}}{6.94} & \RetargetStat{\underline{68.2}}{34.5} & \RetargetStat{\mathbf{41.5}}{21.8} & \RetargetStat{9.32}{9.38} & \RetargetStat{33.7}{46.6} & \RetargetStat{134.3}{197.8} \\
AnyTeleop~\cite{qin2023anyteleop} & \RetargetStat{11.2}{7.99} & \RetargetStat{19.1}{31.8} & \RetargetStat{121.7}{109.3} & \RetargetStat{8.06}{5.76} & \RetargetStat{37.2}{42.8} & \RetargetStat{98.3}{105.4} & \RetargetStat{8.11}{5.35} & \RetargetStat{36.7}{38.8} & \RetargetStat{95.0}{87.9} \\
Contact PyRoki~\cite{kim2025pyroki} & \RetargetStat{12.4}{11.2} & \RetargetStat{33.3}{44.0} & \RetargetStat{210.7}{200.7} & \RetargetStat{15.6}{14.7} & \RetargetStat{25.5}{38.4} & \RetargetStat{226.6}{192.0} & \RetargetStat{\underline{5.76}}{4.42} & \RetargetStat{56.2}{39.2} & \RetargetStat{87.7}{79.5} \\
\midrule
\algabbr{} (Ours) & \RetargetStat{\mathbf{4.10}}{1.76} & \RetargetStat{\mathbf{82.5}}{9.5} & \RetargetStat{\mathbf{47.2}}{23.4} & \RetargetStat{\mathbf{5.01}}{3.38} & \RetargetStat{\mathbf{69.9}}{30.0} & \RetargetStat{\underline{53.0}}{25.2} & \RetargetStat{\mathbf{3.41}}{1.14} & \RetargetStat{\mathbf{91.8}}{3.6} & \RetargetStat{\mathbf{25.1}}{9.3} \\
\bottomrule
\end{tabular}
\endgroup
}
\end{table*}

\begin{table*}[t]
\centering
\caption{Ablation of the contact and object-pose information used by residual RL dynamic retargeting, starting from the same \algabbr{} kinematic references. Same protocol and units as Table~\ref{tab:retargeting_comparison}; bold marks the best mean within each hand and metric.}
\label{tab:ablation}
\resizebox{\textwidth}{!}{%
\begingroup
\setlength{\tabcolsep}{3pt}
\renewcommand{\arraystretch}{1.18}
\begin{tabular}{llccccccccc}
\toprule
 &  & \multicolumn{3}{c}{\textbf{Dex3}} & \multicolumn{3}{c}{\textbf{Allegro}} & \multicolumn{3}{c}{\textbf{Sharpa}} \\
\cmidrule(lr){3-5} \cmidrule(lr){6-8} \cmidrule(lr){9-11}
\textbf{Contact} & \textbf{Object Pose} & C-ADD $\downarrow$ & SR $\uparrow$ & Patch $\downarrow$ & C-ADD $\downarrow$ & SR $\uparrow$ & Patch $\downarrow$ & C-ADD $\downarrow$ & SR $\uparrow$ & Patch $\downarrow$ \\
 &  & ($10^{-2}$\,m) & (\%) & (mm) & ($10^{-2}$\,m) & (\%) & (mm) & ($10^{-2}$\,m) & (\%) & (mm) \\
\midrule
$\times$ & $\checkmark$ & \RetargetStat{6.50}{4.94} & \RetargetStat{51.2}{44.1} & \RetargetStat{91.9}{74.6} & \RetargetStat{8.53}{8.42} & \RetargetStat{47.6}{42.7} & \RetargetStat{122.0}{142.0} & \RetargetStat{6.20}{8.84} & \RetargetStat{76.6}{31.3} & \RetargetStat{74.6}{103.2} \\
$\checkmark$ & $\times$ & \RetargetStat{8.04}{4.21} & \RetargetStat{42.5}{34.0} & \RetargetStat{72.1}{48.3} & \RetargetStat{8.61}{3.65} & \RetargetStat{41.6}{35.7} & \RetargetStat{53.1}{44.0} & \RetargetStat{7.42}{2.27} & \RetargetStat{54.3}{41.3} & \RetargetStat{47.1}{25.7} \\
\midrule
$\checkmark$ & $\checkmark$ & \RetargetStat{\mathbf{4.10}}{1.76} & \RetargetStat{\mathbf{82.5}}{9.5} & \RetargetStat{\mathbf{47.2}}{23.4} & \RetargetStat{\mathbf{5.01}}{3.38} & \RetargetStat{\mathbf{69.9}}{30.0} & \RetargetStat{\mathbf{53.0}}{25.2} & \RetargetStat{\mathbf{3.41}}{1.14} & \RetargetStat{\mathbf{91.8}}{3.6} & \RetargetStat{\mathbf{25.1}}{9.3} \\
\bottomrule
\end{tabular}
\endgroup
}
\end{table*}

\textbf{Object and Contact Ablation.}
To isolate the contribution of contact information in our residual RL formulation, we compare the full method with variants that remove contact or object pose information from the observations, rewards, and termination conditions while keeping the \algabbr{} references fixed.
The results are shown in Table~\ref{tab:ablation}. Using both sources achieves the highest SR and lowest C-ADD across all three hands. Removing contact information reduces SR from $82.5\%$, $69.9\%$, and $91.8\%$ to $51.2\%$, $47.6\%$, and $76.6\%$ for Dex3, Allegro, and Sharpa, respectively; removing object pose information further reduces it to $42.5\%$, $41.6\%$, and $54.3\%$.

The two sources play complementary roles. With object pose information alone, policies better track the object trajectory but deviate substantially from the demonstrated grasp geometry. With contact information alone, policies better preserve the demonstrated interaction but track the object trajectory less accurately and succeed less often. Thus, object pose information guides task-level object motion, while contact information grounds the policy in the demonstrated hand-object interaction; combining both yields the strongest dynamic retargeting performance.

\textbf{Embodiment Analysis.}
Performance differences remain across robot embodiments after dynamic retargeting. Sharpa, whose five fingers and 22 DoF most closely resemble the human hand, achieves the highest SR and lowest final patch distance. Allegro's larger hand and thicker fingers make human-scale grasps more difficult to reproduce and increase the likelihood of table collisions, leading to more frequent early termination. Although Dex3 has only three fingers and 7 DoF, its thinner, more human-scale fingers provide greater clearance for grasps that require operating close to the table, such as the lightbulb, or within narrow object geometry, such as inserting a finger through the torus. As a result, Dex3 achieves higher SR than Allegro despite its lower DoF. These results highlight that improved retargeting reduces, but does not eliminate, constraints imposed by the target robot's morphology.

\subsection{Sim-to-Real Visuomotor Policy Evaluation}
\label{sec:distillation_experiments}
Table~\ref{tab:sharpa_distillation_sr} reports the success rate of the privileged teacher and visuomotor student in simulation, followed by zero-shot deployment of the student on hardware.

\begin{table*}[t]
    \centering
    \caption{Sharpa task success rates (\%). Teacher and student simulation
    use 2,048 and 64 evaluation rollouts per category.
    Hardware uses 30 trials per category across three objects
    (10 poses each). Overall is the average across categories.}
    \label{tab:sharpa_distillation_sr}
    \setlength{\tabcolsep}{3.5pt}
    \renewcommand{\arraystretch}{1.05}
    \small
    \begin{tabular*}{\textwidth}{@{\extracolsep{\fill}}lccccccccccc@{}}
        \toprule
        & Alarm
        & Apple
        & Bowl
        & Cube
        & Cup
        & Flash.
        & Hammer
        & Bulb
        & Torus
        & Wine
        & Overall \\
        \midrule
        Teacher (sim)
        & 90.38 & 93.99 & 88.48 & 92.14 & 92.58
        & 95.85 & 83.84 & 96.19 & 92.19 & 92.19 & 91.78 \\

        Student (sim)
        & 100.00 & 100.00 & 87.50 & 76.56 & 100.00
        & 100.00 & 92.19 & 98.44 & 96.88 & 85.94 & 93.75 \\

        Student (real)
        & 86.67 & 86.67 & 90.00 & 86.67 & 93.33
        & 86.67 & 90.00 & 100.00 & 80.00 & 93.33 & 89.33 \\
        \bottomrule
    \end{tabular*}
\end{table*}

\textbf{Distillation in Simulation.}
We evaluate the teacher and student on unseen simulation rollouts with newly sampled object poses, scales, and physical properties, using 2,048 rollouts per category for the teacher and 64 for the student. The student retains the teacher's performance, achieving a overall average SR of $93.8\%$ compared with $91.8\%$ for the teacher and matching or exceeding it in seven of ten categories. The largest drops occur for the cube ($92.1\%\to76.6\%$) and wineglass ($92.2\%\to85.9\%$). Unlike the teacher, which observes randomized object properties as privileged states, the student must infer their effects from point clouds and proprioception. Additional evaluation details are provided in the Supplemental Material.


\begin{figure}[t]
    \centering
    \includegraphics[width=\linewidth]{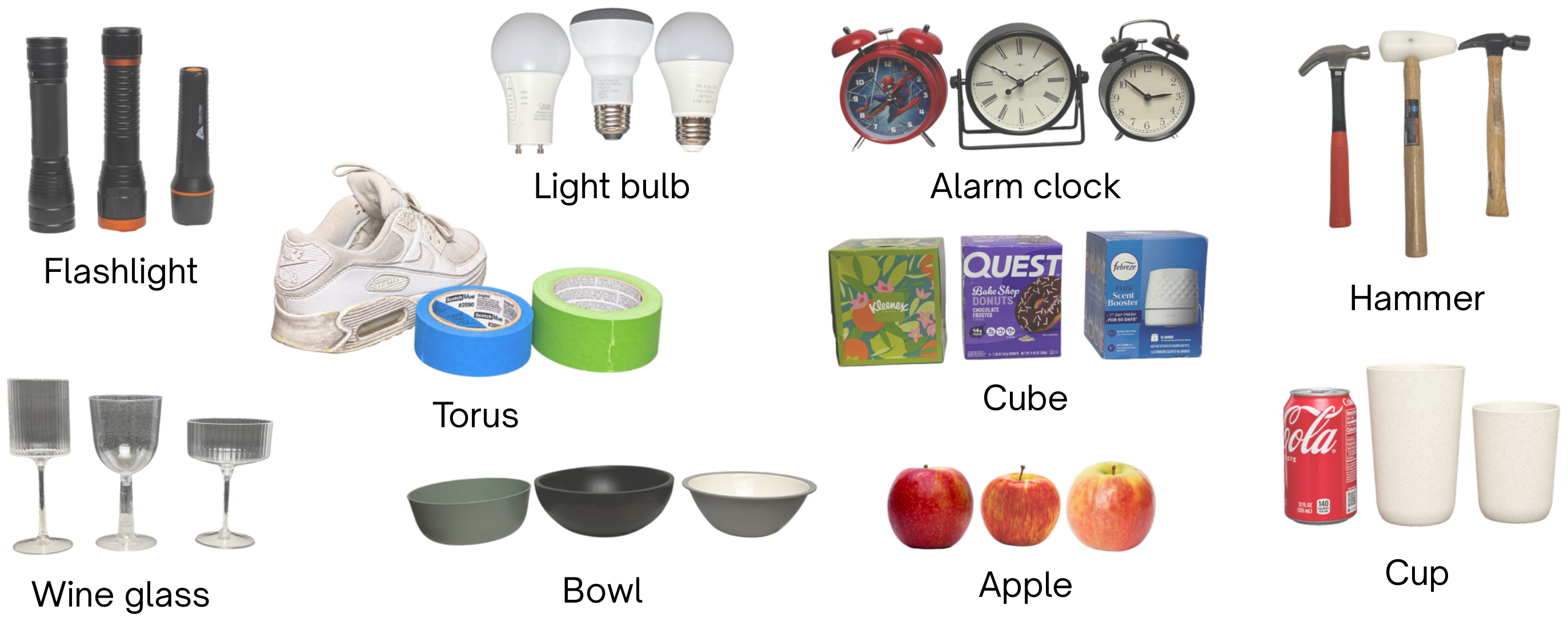}
    \caption{\textbf{Real-World Object Set.} 30 objects used for zero-shot sim-to-real evaluation, comprising three instances from each of the ten object categories.}
    \label{fig:real_world_objects}
\end{figure}

\textbf{Zero-Shot Sim-to-Real.}
Across 300 hardware trials (Figure~\ref{fig:real_experiments}), the visuomotor policies achieve $89.3\%$ zero-shot success, $4.4$ percentage points below the simulated student, without using real-world training data. Every category achieves at least $80\%$ SR, with the lightbulb succeeding in all 30 trials. Collision avoidance also transfers to hardware: the policies successfully grasp objects lying flat on the rigid table, including the hammer, flashlight, and lightbulb, without hard table collisions. The partially observed wineglass achieves $93.3\%$ SR, demonstrating robustness to incomplete point-cloud observations. Note that the transparent wineglass is invisible to the LiDAR camera, so we place ping pong balls in its bowl for partial observability


\textbf{Failure Modes.}
We observed no failures due to hard collisions with the table. Instead, failures primarily arose from two other forms of inaccurate spatial reasoning. First, object localization errors were strongly dependent on the object's position within the randomized $10\,\text{cm}\times10$\,cm region; for some object categories, failures were concentrated entirely within particular regions of the workspace. To further characterize these pose-dependent failure patterns, we provide visualizations of the distribution of successes and failures across the 30 trials for each object category in the Supplemental Material. Second, we observed errors in interpreting object geometry. Unlike the pose-dependent localization failures, the hand approached the correct object location but formed an inadequate pre-grasp, typically by not opening sufficiently to accommodate the object's geometry. As the hand subsequently attempted to establish contact from the side, it instead pushed the object away. This failure occurred for the red alarm clock, the thickest of the three physical instances (Figure~\ref{fig:real_world_objects}), but was not observed for the other two instances.
The torus has the lowest real-world SR ($80.0\%$), substantially below both teacher and student performance in simulation. Two of the three physical instances are rolls of tape (Figure~\ref{fig:real_world_objects}), which are substantially thinner than the bagel-shaped GRAB reference. This geometric variation is not captured by our anisotropic scale randomization, making these instances particularly out of distribution. In contrast, the cube performs substantially better in the real world than in student simulation. The three physical cubes (Figure~\ref{fig:real_world_objects}) have nearly identical geometry, material properties, and approximately uniform mass distributions, differing primarily in mass. We conjecture that this improvement is due to the narrower object variation represented by the physical cube instances relative to the simulation randomization.

\section{Conclusion}
We present \textit{\algname{}}, a framework that transforms reconstructed human hand-object interactions into zero-shot sim-to-real visuomotor policies. Our framework combines \algabbr{} for morphology- and contact-aware kinematic retargeting, residual RL that leverages object pose and contact information for dynamic retargeting, and point-cloud policy distillation for real-world deployment.

Our experiments across three-, four-, and five-fingered hands lead to three main conclusions. First, explicitly accounting for morphology and contact during kinematic retargeting produces more faithful references: \algabbr{} achieves the highest contact F1 and lowest patch distance across all three hands, which in turn improves downstream task success, object tracking, and contact preservation under the same residual RL formulation. Second, object pose and contact information play complementary roles in dynamic retargeting: object pose guides trajectory tracking and task completion, while contact information promotes preservation of the demonstrated interaction, and combining both yields the strongest performance. Third, our resulting sim-to-real visuomotor policies transfer effectively, achieving $89.3\%$ zero-shot success while avoiding hard collisions with the rigid table. The ranking among hands follows the embodiment gap to the human hand, while \algabbr{} improves over the strongest baseline for each hand using the same hyperparameters throughout.


Several limitations motivate future work. We currently train separate residual RL and visuomotor policies for each object category and evaluate hardware deployment on a single arm-hand system. Distilling demonstrations across categories into a unified multi-task policy and deploying on the other hands are natural next steps. 
Furthermore, our demonstrations come from ten motion capture trajectories, whereas the scalability argument for human motion data rests on reconstructions from monocular video; since \algabbr{} operates on hand and object meshes and runs at 180 frames per second, extending the framework to such reconstructions, including bimanual interactions, is a promising direction.
Finally, improving generalization to a broader range of object poses and shapes in a computationally and data-efficient manner remains an important direction for future work.

\bibliographystyle{IEEEtran}
\bibliography{refs}

\clearpage
\appendices
\section{Contact evaluation protocol}
\label{app:retarget-metrics}
\ifdefined\arxivversion
This protocol scores the kinematic retargeting results in
Section~\ref{sec:kinematic_experiments} and, applied to the final state
of each rollout, the dynamic retargeting results in
Section~\ref{sec:dynamic_experiments}.
\else
This protocol scores the kinematic retargeting results in
Appendix~\ref{app:kinematic} and, applied to the final state of each
rollout, the dynamic retargeting results in Appendix~\ref{app:rl-contact}.
\fi

We use $m$ for a retargeting attempt, $t$ for a retained frame, $h$ for
a human hand part, $k$ for a robot hand part, and $v$ for an object
vertex. A part is a finger or the palm. Unless aggregation across
attempts is being discussed, the index $m$ is suppressed. Distances
are computed in meters and converted to millimeters for reporting;
$\tau_H,\tau_B$ are contact tolerances. In particular, $B$ denotes a binary robot
contact indicator, not a robot surface or a blend-shape matrix.

\subsection{Trajectory alignment and geometry}
Section~\ref{sec:baselines} lists the compared methods.
For Ours, OmniRetarget, and Contact PyRoki, we retain the existing robot
configurations and hand geometry. We use \texttt{joints\_pk\_raw}, reordered
by \texttt{idx\_pk2sapien}; interpolated object/contact arrays are reduced
to their original endpoints using the recorded interpolation counts:
each original row is recovered by offsetting its index by the number of
rows inserted up to and including it. We remove the synthetic HOME row from these outputs. All six methods and
the MANO reference therefore use exactly the same original demonstration
frames, with human indices $[0,1,\ldots,N_{\rm demo}-1]$ and no interpolated
or synthetic frames. Each retained frame has equal weight; reported frame
fractions are not elapsed-time fractions. Object poses and human patch
support match the previous raw evaluation after this common frame selection.
Contact predictions are recomputed from geometry, independently of stored
robot contact flags.

Robot surfaces are obtained from URDF forward kinematics and collision
geometry, using convex hulls for mesh collision shapes. Human geometry is
the posed MANO surface. Patches are compared on a common object mesh in
object coordinates, using the aligned object poses. Specifically, let
$T^H_{O,t}$ and $T^B_{O,t}$ be the object-to-world transforms in the
human demonstration and robot trajectory, respectively, and let
$T_{b,t}(q_t)$ be the transform from robot link $b$ to the world, from
forward kinematics at robot configuration $q_t$. A human world point
$y_w^H$ and a point $y_b^B$ in the frame of link $b$ are transformed as
\begin{equation}
\begin{aligned}
\widetilde y^H&=(T^H_{O,t})^{-1}\widetilde y_w^H,\\
\widetilde y^B&=(T^B_{O,t})^{-1}T_{b,t}(q_t)\widetilde y_b^B,
\end{aligned}
\end{equation}
where tildes denote homogeneous coordinates and each $T$ is a rigid
$4\times4$ transform. This compares locations relative to the object
even when its world pose differs between demonstration and retargeting.
The evaluator samples
robot collision surfaces at nominal spacings of $2\,\mathrm{mm}$ for
hands, and also queries object vertices
against robot shapes. These are discretized geometric measurements,
not exact continuous collision certificates. Signed queries depend on
mesh orientation and closure; the benchmark includes an open bowl mesh.

The hand-part set includes the palm and individual fingers. Human parts
are mapped to the robot's available parts by a fixed map $g$. Allegro
merges human middle and ring into its middle finger; Dex3 maps index and
middle to its index finger, and ring and pinky to its middle finger.
Sharpa uses one-to-one correspondence. Palm and thumb retain their
identities. Many-to-one human supports are unioned before scoring, so a
robot part is not counted twice for the same frame or surface vertex.

\paragraph{Human part labels}
The anatomical labels are fixed before evaluating any method. Each MANO
vertex is assigned the joint with the largest skinning weight. A triangle
receives the majority label of its three vertices; when all three labels
differ, the first vertex breaks the tie. Joint labels are then grouped
into fingers and palm. The label of the closest triangle therefore
identifies $h^*_{tv}$. This construction uses the original demonstrated
MANO geometry, not the morphology-optimized hand produced by our method.

\paragraph{Signed geometric queries}
For a solid $\Omega$, write $\phi_\Omega(x)$ for signed distance to its boundary:
\begin{equation}
\label{eq:retarget-sdf}
\phi_\Omega(x)=
\begin{cases}
-\operatorname{dist}(x,\partial\Omega),&x\text{ inside }\Omega,\\
\phantom{-}\operatorname{dist}(x,\partial\Omega),&x\text{ outside }\Omega,
\end{cases}
\end{equation}
with zero on the boundary. Write $\phi_O$ for the corresponding object
query. For the open bowl mesh, the implementation uses a generalized
winding-number sign; an exact inside/outside interpretation is not
available for an open surface.

Let $\mathcal X^B_{tk}$ be the robot surface samples belonging to part
$k$ at frame $t$, expressed in object coordinates, let $\mathcal R_k(t)$
be the collision shapes of part $k$, and let $\mathcal V_t(\Omega)$ index
the object vertices queried against shape $\Omega$.
The robot part--object queries comprise both directions:
\begin{equation}
\label{eq:retarget-directed-queries}
\begin{aligned}
\mathcal D^k_t
={}&\{\phi_O(y):y\in\mathcal X^B_{tk}\}\\
&{}\cup\!\!\bigcup_{\Omega\in\mathcal R_k(t)}
       \{\phi_\Omega(x_v):v\in\mathcal V_t(\Omega)\}.
\end{aligned}
\end{equation}
The reverse queries use object vertices within an expanded link bounding
box and require collision shapes that support signed queries.
The bounding-box margin is at least $10\,\mathrm{mm}$ and therefore
includes vertices that could satisfy any tested proximity threshold.

\subsection{Location-unaware contact}
Let $\mathcal X^H_{th}$ contain the posed MANO vertices labeled as human
part $h$. The evaluated signed part--object distances are
\begin{equation}
\label{eq:retarget-part-distances}
s^H_{th}=\min_{y\in\mathcal X^H_{th}}\phi_O(y),
\qquad
s^B_{tk}=\min_{\zeta\in\mathcal D^k_t}\zeta.
\end{equation}
The desired and predicted frame--part contact indicators are
\begin{equation}
\begin{split}
H^{\rm LU}_{tk}(\tau)
 &=\mathbf{1}\!\left[\min_{h:g(h)=k}s^H_{th}\leq\tau\right],\\
B^{\rm LU}_{tk}(\tau)
 &=\mathbf{1}[s^B_{tk}\leq\tau].
\end{split}
\end{equation}
Thus a part inside the object counts as in contact. The counts are
\begin{equation}
\label{eq:retarget-weighted-counts}
\begin{split}
\mathrm{TP}&=\textstyle\sum_{t,k}H^{\rm LU}_{tk}B^{\rm LU}_{tk},\\
\mathrm{FP}&=\textstyle\sum_{t,k}(1-H^{\rm LU}_{tk})B^{\rm LU}_{tk},\\
\mathrm{FN}&=\textstyle\sum_{t,k}H^{\rm LU}_{tk}(1-B^{\rm LU}_{tk}).
\end{split}
\end{equation}
These counts include all evaluated frames, including approach and release
frames without human contact, so undesired robot contacts contribute
false positives. No temporal tolerance is used.

\subsection{Location-aware contact}
This is the family used in the main table. Its indicators
$H^{\rm LA}_{tkv}$ and $B^{\rm LA}_{tkv}$ are the main text's
$H_{tkv}$ and $B_{tkv}$, with the human and robot tolerances written
separately here.
Let $x_v$ be object vertex $v$ and assign it the surface-area weight
\begin{equation}
a_v=\frac{1}{3}\sum_{f\ni v}\operatorname{area}(f).
\end{equation}
Let $d^H_{tv}$ be its unsigned distance to the entire posed human hand
surface, and let $h^*_{tv}$ be the human part assigned to the closest
MANO triangle. Explicitly, for the posed MANO surface $\mathcal S_t^H$,
\begin{equation}
d^H_{tv}=\min_{y\in\mathcal S_t^H}\|x_v-y\|_2.
\end{equation}
The nearest triangle is found on the entire human hand before its
label is mapped by $g$; we do not independently search each human finger.
For the collision shapes $\mathcal R_k(t)$ of robot part $k$, define
\begin{equation}
d^B_{tkv}=\min_{\Omega\in\mathcal R_k(t)}
          \operatorname{dist}(x_v,\partial\Omega).
\end{equation}
This is the minimum unsigned distance to the component surfaces; it is
computed as the minimum of the absolute component signed distances,
not the absolute value of their minimum.
For human and robot tolerances $\tau_H$ and $\tau_B$, define
\begin{equation}
\begin{split}
H^{\rm LA}_{tkv}(\tau_H)
 &=\mathbf{1}[d^H_{tv}\leq\tau_H]\,
   \mathbf{1}[g(h^*_{tv})=k],\\
B^{\rm LA}_{tkv}(\tau_B)
 &=\mathbf{1}[d^B_{tkv}\leq\tau_B].
\end{split}
\end{equation}
The counts are those of Eq.~\eqref{eq:retarget-contact-counts} in the main
text, which sum over $(t,k,v)$ with weight $a_v$. The resulting masses have units of surface area accumulated
over frames. Human and robot contact must agree in frame, mapped part,
and object vertex to contribute a true positive. Robot patches are
queried across the object, not just on the human patch, so robot-only
regions and frames contribute false positives. Spatial resolution is
limited by the object mesh and the collision geometry.

The primary tables use matched thresholds $\tau_H=\tau_B=\tau$ for
$\tau\in\{1,5,10\}\,\mathrm{mm}$. This changes both the desired and
predicted sets; neither precision nor recall is required to be monotonic.
For example, if the robot index finger touches a different side of the
object from the demonstrated index-finger patch, the missing desired
region contributes FN and the extra robot region contributes FP.
The location-unaware indicators can nevertheless both be one. Because
location-aware contact uses unsigned surface distance, a vertex deep
inside a robot shape need not count as contact.

\subsection{Precision, recall, F1, and failure handling}
\label{app:retarget-aggregation}
We pool counts or area within each trajectory before computing
\begin{equation}
\begin{aligned}
\mathrm{Prec}&=\frac{\mathrm{TP}}{\mathrm{TP}+\mathrm{FP}},\qquad
\mathrm{Rec}=\frac{\mathrm{TP}}{\mathrm{TP}+\mathrm{FN}},\\
\mathrm{F1}&=\frac{2\mathrm{TP}}{2\mathrm{TP}+\mathrm{FP}+\mathrm{FN}}.
\end{aligned}
\end{equation}
All reference sequences in the present benchmark have positive desired
support at all three thresholds. If a successful trajectory predicts no
contacts, ordinary precision is undefined because
$\mathrm{TP}+\mathrm{FP}=0$; the benchmark score assigns precision zero
in this case. Recall and F1 are also zero because desired support is
positive. Contact-free reference sequences, if added in a
future benchmark, require a separately declared absence-detection
protocol and must not silently receive perfect contact-preservation
scores.

For every method, a confirmed failed retargeting receives zero in the
failure-adjusted contact scores; successful runs use the convention
above. Equivalently, for a per-attempt score $\xi_m$ (precision,
recall, F1, or coverage), the failure-adjusted mean $\overline\xi_{\rm FA}$
sums $\xi_m$ over the method's successful attempts and divides by the
attempted count $N_{\rm att}=10$ per hand. With $N_{\rm succ}$ successful
attempts, the failure rate is
\begin{equation}
f_{\rm fail}=\frac{N_{\rm att}-N_{\rm succ}}{N_{\rm att}}.
\end{equation}
Unattempted runs and evaluator errors are not reclassified as
retargeting failures. A missing trajectory has no measured contact
support or distance; assigning zero to a benchmark utility
does not fabricate these quantities.

Tables report the mean and population standard deviation of the
per-attempt scores, including failure zeros:
\begin{equation}
\sigma_\xi=\sqrt{\frac{1}{N_{\rm att}}\sum_m
  (\widetilde\xi_m-\overline\xi_{\rm FA})^2},
\end{equation}
where $\widetilde\xi_m$ equals $\xi_m$ for a successful attempt and zero
for a failed one.
F1 is averaged after computing each trajectory's F1; it is not the
harmonic mean of the displayed macro precision and recall. Standard
deviations describe variation between demonstrations, not uncertainty
across independent training seeds or confidence intervals.

\paragraph{Matched sequences}
Let $\mathcal I$ be the intersection of successful sequences across all
methods for a given hand; $|\mathcal I|$ is 9, 8, and 10 for Allegro,
Dex3, and Sharpa, respectively. Comparisons on $\mathcal I$ use the same
inputs for every method and carry no failure penalty. Pooled (micro)
scores on $\mathcal I$ sum TP, FP, and FN over its trajectories before
computing precision, recall, and F1, and therefore weight trajectories by
contact support.

\subsection{Choice of contact tolerance}
The nominal tolerance is $5\,\mathrm{mm}$, with $1\,\mathrm{mm}$ and
$10\,\mathrm{mm}$ probing stricter and more permissive geometric
agreement. This also retains the fixed $5\,\mathrm{mm}$ human patch used
for the continuous-distance diagnostic. The thresholds are operational
definitions of geometric contact; they are not estimates of annotation
accuracy or a physical contact-sensor noise level.

An audit of the ten object meshes gives per-object median edge lengths
between $0.88$ and $2.10\,\mathrm{mm}$. At $1\,\mathrm{mm}$, the median
nonempty human patch contains approximately 200 vertices per frame and
mapped part, so strict patches are not generally single-vertex events.
However, area is still integrated at mesh vertices. A tolerance near the
mesh spacing motivates a surface-resampling convergence study before
claiming physically calibrated millimeter accuracy. Robot-patch queries
measure point-to-shape surface distance directly; the $2\,\mathrm{mm}$
robot sampling interval used in other collision computations does not
quantize the location-aware distances to $2\,\mathrm{mm}$ increments.
No claim that $5\,\mathrm{mm}$ is an optimal noise-calibrated tolerance
is made. The relative performance across all three thresholds, together
with continuous patch distance, is the relevant robustness evidence.

\subsection{Continuous error and coverage}
For a human patch defined at $\tau_H$, let
$A_H=\sum_{t,k,v}a_vH^{\rm LA}_{tkv}(\tau_H)$. The mean patch-to-part
distance and coverage at robot tolerance $\tau_B$ are
\begin{equation}
\begin{split}
D_{\rm patch}
 &=\frac{\sum_{t,k,v}a_vH^{\rm LA}_{tkv}d^B_{tkv}}{A_H},\\
\mathrm{Cov}(\tau_B)
 &=\frac{\sum_{t,k,v}a_vH^{\rm LA}_{tkv}
                  \mathbf{1}[d^B_{tkv}\leq\tau_B]}{A_H}.
\end{split}
\end{equation}
Every desired vertex contributes to the mean even if the robot makes no
contact there. With the same patch definition, coverage is exactly
location-aware recall, so we do not treat it as independent evidence.
The area-weighted 95th-percentile distance is
\begin{equation}
D_{95}=\inf\!\left\{\zeta:
\frac{\sum_{t,k,v}a_vH^{\rm LA}_{tkv}
                   \mathbf{1}[d^B_{tkv}\leq\zeta]}{A_H}\geq0.95\right\}.
\end{equation}
We use $\tau_H=5\,\mathrm{mm}$ for the primary continuous-distance
comparison. Its mean and percentile do not depend on the robot coverage
tolerance. These are means of per-trajectory statistics, not a single
percentile over the entire dataset. Geometry tables use $\mathcal I$ for
all methods; failure cases remain N/A.

Arm--object, terrain, self-collision, and slip scores are outside this
hand--object comparison and are not imputed for floating-hand outputs.
Geometry for a failed retargeting remains undefined.

Contact scores alone do not establish dynamic feasibility or downstream
task success. The floating-hand references require a separately validated
arm-placement stage if used in an arm-equipped environment.

\section{Kinematic retargeting}
\label{app:kinematic}
\subsection{Additional quantitative results}
\label{app:kinematic-results}
Table~\ref{tab:retarget-sweep} gives location-aware precision, recall, and F1 at all three matched tolerances. Table~\ref{tab:retarget-geometry-detail} adds the patch-distance tail on the matched sequences.

Alternative aggregations leave the F1 comparison unchanged. OmniRetarget is the only method with retargeting failures; averaging over its successful runs alone raises its $5\,\mathrm{mm}$ location-aware F1 from $12.8\%$ to $14.3\%$ on Allegro and from $7.6\%$ to $9.6\%$ on Dex3. On the matched sequences $\mathcal I$, \algabbr{} has the highest macro location-aware F1 at every tolerance on every hand. Pooled (micro) scores at $5\,\mathrm{mm}$ on $\mathcal I$ give the same result: \algabbr{} has the highest location-aware F1 ($28.8\%$, $42.3\%$, and $39.4\%$ for Allegro, Dex3, and Sharpa), while OmniRetarget has the highest or second-highest pooled precision on every hand but the lowest recall. Ordinary precision, which excludes failed runs and empty predictions instead of scoring them as zero, differs from failure-adjusted precision only for methods with such runs; per-sequence values are included in the numerical release.

\begin{table*}[t]
\centering
\footnotesize
\caption{Location-aware contact across matched tolerances $\tau_H=\tau_B=\tau$. Failure-adjusted macro percentages (mean $\pm$ population SD) over ten attempts per hand and method; coverage equals recall. The $5\,\mathrm{mm}$ F1 repeats Table~\ref{tab:retarget-main}, and location-unaware F1 is shown in Fig.~\ref{fig:retarget-threshold}. Bold and underlining mark the highest and second-highest distinct unrounded means within each hand, tolerance, and metric; ties share formatting.}
\label{tab:retarget-sweep}
\begingroup
\setlength{\tabcolsep}{1pt}
\renewcommand{\arraystretch}{1.13}
\begin{tabular*}{\textwidth}{@{\extracolsep{\fill}}lrrrrrrrrr@{}}
\toprule
 & \multicolumn{3}{c}{$\tau=1\,\mathrm{mm}$} & \multicolumn{3}{c}{$\tau=5\,\mathrm{mm}$} & \multicolumn{3}{c}{$\tau=10\,\mathrm{mm}$} \\
\cmidrule(lr){2-4} \cmidrule(lr){5-7} \cmidrule(lr){8-10}
Method & Prec.\ $\uparrow$ & Rec.\ $\uparrow$ & F1 $\uparrow$ & Prec.\ $\uparrow$ & Rec.\ $\uparrow$ & F1 $\uparrow$ & Prec.\ $\uparrow$ & Rec.\ $\uparrow$ & F1 $\uparrow$ \\
\midrule
\multicolumn{10}{@{}l}{\textit{Allegro}} \\
OmniRetarget & \RetargetStat{\mathbf{9.4}}{12.3} & \RetargetStat{1.1}{1.9} & \RetargetStat{1.8}{3.1} & \RetargetStat{\mathbf{35.6}}{31.1} & \RetargetStat{8.5}{11.4} & \RetargetStat{12.8}{16.0} & \RetargetStat{\mathbf{53.1}}{30.6} & \RetargetStat{18.0}{16.8} & \RetargetStat{25.2}{20.8} \\
DexPilot & \RetargetStat{2.3}{1.6} & \RetargetStat{1.9}{1.5} & \RetargetStat{2.0}{1.5} & \RetargetStat{15.3}{7.7} & \RetargetStat{12.5}{7.3} & \RetargetStat{13.1}{6.8} & \RetargetStat{26.8}{11.2} & \RetargetStat{24.4}{10.3} & \RetargetStat{24.3}{9.4} \\
\texttt{Position} & \RetargetStat{5.3}{3.3} & \RetargetStat{\underline{5.6}}{7.0} & \RetargetStat{4.4}{3.2} & \RetargetStat{26.6}{11.8} & \RetargetStat{\underline{23.1}}{18.5} & \RetargetStat{\underline{22.0}}{10.8} & \RetargetStat{40.5}{12.9} & \RetargetStat{\underline{37.2}}{19.7} & \RetargetStat{\underline{36.0}}{11.8} \\
AnyTeleop & \RetargetStat{2.2}{1.7} & \RetargetStat{1.9}{1.7} & \RetargetStat{2.0}{1.7} & \RetargetStat{15.4}{7.5} & \RetargetStat{12.5}{7.7} & \RetargetStat{13.2}{7.1} & \RetargetStat{26.4}{10.9} & \RetargetStat{24.0}{10.8} & \RetargetStat{24.0}{9.5} \\
Contact PyRoki & \RetargetStat{\underline{7.6}}{5.6} & \RetargetStat{3.9}{3.5} & \RetargetStat{\underline{4.5}}{3.9} & \RetargetStat{24.9}{16.0} & \RetargetStat{13.2}{10.5} & \RetargetStat{15.4}{11.0} & \RetargetStat{35.7}{20.1} & \RetargetStat{22.1}{13.4} & \RetargetStat{25.1}{13.0} \\
Ours & \RetargetStat{7.5}{3.7} & \RetargetStat{\mathbf{9.3}}{4.5} & \RetargetStat{\mathbf{8.0}}{3.8} & \RetargetStat{\underline{29.0}}{7.4} & \RetargetStat{\mathbf{33.5}}{11.5} & \RetargetStat{\mathbf{30.3}}{7.4} & \RetargetStat{\underline{44.2}}{8.2} & \RetargetStat{\mathbf{50.5}}{11.5} & \RetargetStat{\mathbf{46.4}}{7.5} \\
\midrule
\multicolumn{10}{@{}l}{\textit{Unitree Dex3}} \\
OmniRetarget & \RetargetStat{\underline{2.5}}{5.1} & \RetargetStat{0.7}{1.5} & \RetargetStat{1.1}{2.4} & \RetargetStat{\underline{22.6}}{23.5} & \RetargetStat{5.5}{8.4} & \RetargetStat{7.6}{11.3} & \RetargetStat{\underline{36.6}}{28.2} & \RetargetStat{12.5}{13.7} & \RetargetStat{16.9}{17.0} \\
DexPilot & \RetargetStat{0.7}{1.9} & \RetargetStat{0.3}{0.7} & \RetargetStat{0.4}{1.0} & \RetargetStat{3.7}{7.0} & \RetargetStat{2.0}{2.8} & \RetargetStat{2.5}{4.0} & \RetargetStat{10.1}{11.7} & \RetargetStat{6.8}{6.2} & \RetargetStat{7.9}{7.9} \\
\texttt{Position} & \RetargetStat{2.2}{2.0} & \RetargetStat{\underline{3.8}}{4.8} & \RetargetStat{\underline{2.2}}{1.9} & \RetargetStat{9.9}{8.1} & \RetargetStat{\underline{13.8}}{10.4} & \RetargetStat{\underline{10.2}}{7.6} & \RetargetStat{17.0}{11.6} & \RetargetStat{\underline{24.7}}{15.4} & \RetargetStat{\underline{18.5}}{11.3} \\
AnyTeleop & \RetargetStat{0.7}{1.9} & \RetargetStat{0.3}{0.7} & \RetargetStat{0.4}{1.0} & \RetargetStat{4.0}{6.9} & \RetargetStat{2.2}{2.7} & \RetargetStat{2.7}{3.9} & \RetargetStat{10.4}{11.6} & \RetargetStat{7.0}{6.1} & \RetargetStat{8.1}{7.8} \\
Contact PyRoki & \RetargetStat{1.6}{3.5} & \RetargetStat{3.2}{7.9} & \RetargetStat{2.1}{4.9} & \RetargetStat{7.6}{9.4} & \RetargetStat{13.3}{19.8} & \RetargetStat{9.6}{12.8} & \RetargetStat{15.1}{10.3} & \RetargetStat{22.9}{17.0} & \RetargetStat{18.1}{12.7} \\
Ours & \RetargetStat{\mathbf{8.1}}{3.7} & \RetargetStat{\mathbf{11.5}}{2.6} & \RetargetStat{\mathbf{9.1}}{3.2} & \RetargetStat{\mathbf{33.9}}{10.3} & \RetargetStat{\mathbf{46.0}}{7.3} & \RetargetStat{\mathbf{38.0}}{8.9} & \RetargetStat{\mathbf{50.0}}{11.9} & \RetargetStat{\mathbf{62.6}}{8.4} & \RetargetStat{\mathbf{54.7}}{9.8} \\
\midrule
\multicolumn{10}{@{}l}{\textit{Sharpa Wave}} \\
OmniRetarget & \RetargetStat{9.0}{14.6} & \RetargetStat{2.2}{4.2} & \RetargetStat{3.4}{6.4} & \RetargetStat{\mathbf{61.3}}{34.2} & \RetargetStat{9.5}{12.3} & \RetargetStat{14.5}{18.1} & \RetargetStat{\mathbf{66.9}}{34.0} & \RetargetStat{21.3}{18.0} & \RetargetStat{30.1}{22.5} \\
DexPilot & \RetargetStat{5.7}{3.9} & \RetargetStat{\underline{5.3}}{3.6} & \RetargetStat{5.0}{3.1} & \RetargetStat{24.6}{17.1} & \RetargetStat{\underline{24.5}}{11.2} & \RetargetStat{21.9}{9.8} & \RetargetStat{36.5}{15.6} & \RetargetStat{\underline{44.3}}{17.7} & \RetargetStat{36.9}{13.5} \\
\texttt{Position} & \RetargetStat{6.7}{5.2} & \RetargetStat{5.2}{5.5} & \RetargetStat{5.4}{4.9} & \RetargetStat{34.9}{18.0} & \RetargetStat{20.9}{18.0} & \RetargetStat{23.9}{17.7} & \RetargetStat{54.6}{14.0} & \RetargetStat{30.7}{21.3} & \RetargetStat{34.9}{20.1} \\
AnyTeleop & \RetargetStat{5.5}{4.0} & \RetargetStat{4.7}{3.3} & \RetargetStat{4.6}{3.0} & \RetargetStat{24.0}{16.6} & \RetargetStat{22.7}{10.9} & \RetargetStat{21.0}{9.8} & \RetargetStat{36.5}{15.1} & \RetargetStat{42.7}{16.9} & \RetargetStat{36.3}{13.2} \\
Contact PyRoki & \RetargetStat{\underline{12.3}}{9.1} & \RetargetStat{4.5}{4.1} & \RetargetStat{\underline{6.1}}{5.1} & \RetargetStat{48.5}{17.8} & \RetargetStat{20.8}{15.6} & \RetargetStat{\underline{26.7}}{16.1} & \RetargetStat{\underline{64.8}}{13.0} & \RetargetStat{36.5}{15.5} & \RetargetStat{\underline{44.0}}{13.2} \\
Ours & \RetargetStat{\mathbf{13.1}}{8.1} & \RetargetStat{\mathbf{7.4}}{4.5} & \RetargetStat{\mathbf{9.1}}{5.4} & \RetargetStat{\underline{51.2}}{22.2} & \RetargetStat{\mathbf{31.2}}{15.0} & \RetargetStat{\mathbf{37.2}}{16.6} & \RetargetStat{59.7}{23.6} & \RetargetStat{\mathbf{49.3}}{19.2} & \RetargetStat{\mathbf{52.9}}{19.8} \\
\bottomrule
\end{tabular*}
\endgroup
\end{table*}
\begin{table}[t]
\centering
\footnotesize
\caption{Contact patch distance on matched demonstration sequences with valid retargeting outputs from every method. The human patch uses $5\,\mathrm{mm}$. Entries are mean $\pm$ population SD of sequence statistics. Failed runs have no geometric measurement. Bold and underlining mark the lowest and second-lowest distinct unrounded means within each hand and column; ties share formatting.}
\label{tab:retarget-geometry-detail}
\begingroup
\setlength{\tabcolsep}{2pt}
\renewcommand{\arraystretch}{1.13}
\begin{tabular*}{\columnwidth}{@{\extracolsep{\fill}}lrr@{}}
\toprule
Method & $D_{\rm patch}$ (mm) $\downarrow$ & $D_{95}$ (mm) $\downarrow$ \\
\midrule
\multicolumn{3}{@{}l}{\textit{Allegro ($n_{\mathcal I}=9$)}} \\
OmniRetarget & \RetargetStat{21.2}{13.1} & \RetargetStat{39.2}{22.7} \\
DexPilot & \RetargetStat{\underline{18.7}}{4.9} & \RetargetStat{\underline{38.2}}{7.6} \\
\texttt{Position} & \RetargetStat{26.3}{39.4} & \RetargetStat{52.4}{69.1} \\
AnyTeleop & \RetargetStat{19.0}{5.1} & \RetargetStat{38.8}{7.8} \\
Contact PyRoki & \RetargetStat{21.5}{6.1} & \RetargetStat{43.7}{14.8} \\
Ours & \RetargetStat{\mathbf{9.1}}{2.0} & \RetargetStat{\mathbf{21.0}}{3.7} \\
\midrule
\multicolumn{3}{@{}l}{\textit{Unitree Dex3 ($n_{\mathcal I}=8$)}} \\
OmniRetarget & \RetargetStat{26.9}{10.9} & \RetargetStat{63.0}{16.9} \\
DexPilot & \RetargetStat{42.6}{7.4} & \RetargetStat{62.1}{5.8} \\
\texttt{Position} & \RetargetStat{\underline{18.9}}{6.4} & \RetargetStat{\underline{46.1}}{14.1} \\
AnyTeleop & \RetargetStat{42.1}{7.2} & \RetargetStat{61.7}{5.1} \\
Contact PyRoki & \RetargetStat{26.6}{10.3} & \RetargetStat{62.1}{18.7} \\
Ours & \RetargetStat{\mathbf{7.5}}{1.4} & \RetargetStat{\mathbf{21.2}}{7.5} \\
\midrule
\multicolumn{3}{@{}l}{\textit{Sharpa Wave ($n_{\mathcal I}=10$)}} \\
OmniRetarget & \RetargetStat{21.1}{15.5} & \RetargetStat{39.5}{27.1} \\
DexPilot & \RetargetStat{12.6}{3.7} & \RetargetStat{28.7}{5.5} \\
\texttt{Position} & \RetargetStat{25.4}{39.0} & \RetargetStat{47.4}{67.5} \\
AnyTeleop & \RetargetStat{13.3}{3.7} & \RetargetStat{30.1}{5.2} \\
Contact PyRoki & \RetargetStat{\underline{10.7}}{2.7} & \RetargetStat{\mathbf{20.9}}{5.7} \\
Ours & \RetargetStat{\mathbf{9.7}}{6.2} & \RetargetStat{\underline{21.7}}{11.7} \\
\bottomrule
\end{tabular*}
\endgroup
\end{table}

\subsection{Qualitative comparison with all baselines}
\label{app:kinematic-qualitative}
\begin{figure*}[t]
\centering
\includegraphics[width=\textwidth]{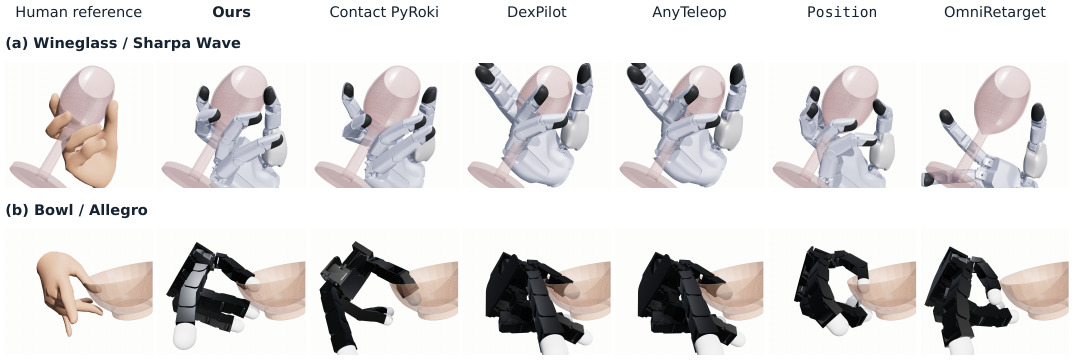}
\caption{\textbf{Qualitative kinematic retargeting with all methods} at the frames of Fig.~\ref{fig:kinematic_comparison}: (a) wineglass with Sharpa and (b) bowl with Allegro.}
\label{fig:kinematic_comparison_full}
\end{figure*}

Figure~\ref{fig:kinematic_comparison_full} extends Fig.~\ref{fig:kinematic_comparison} to all six methods at the same demonstration frames.

\subsection{Threshold and paired-difference diagnostics}
\begin{figure*}[t]
\centering
\includegraphics[width=\textwidth]{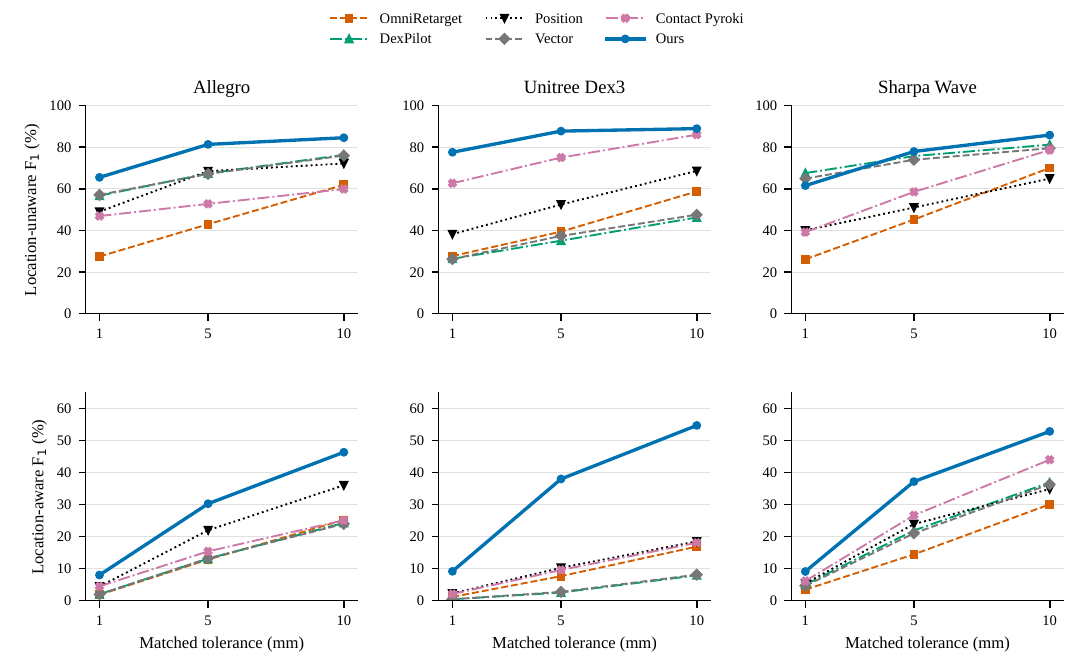}
\caption{Failure-adjusted macro F1 across matched human and robot tolerances.
Every point averages ten attempts per hand and method. Both reference and
predicted supports change with tolerance. All three tolerances are disclosed.}
\label{fig:retarget-threshold}
\end{figure*}

Figure~\ref{fig:retarget-threshold} shows the threshold sweep for both contact
families. Lines connect evaluated
operating points and do not imply measurements at intermediate tolerances.
Table~\ref{tab:retarget-paired-intervals} reports paired differences in
location-aware F1 at 5 mm between Ours and each baseline. We resample
demonstrations, keeping each method's score paired with Ours on the same
input, with 20,000 bootstrap draws and seed 20260917. At the nominal 5\%
level, Ours improves significantly in 13 of the 15 hand--baseline
comparisons: every interval excludes zero on Allegro and Unitree Dex3, and
on Sharpa Wave every interval excludes zero except those against
\texttt{Position} ($[-4.5,\,28.3]$) and Contact PyRoki ($[-7.1,\,26.6]$).
The intervals are pointwise 95\% percentile intervals without a
multiplicity adjustment across the 15 comparisons, so they do not support a
joint claim that Ours outperforms every baseline. Because demonstrations
are the resampling unit, the intervals characterize variation across the
ten selected demonstrations per hand. They should not be interpreted as
population-wide guarantees or variation across policy seeds.
\begin{table}[t]
\centering
\footnotesize
\caption{Paired differences in location-aware F1 at $5\,\mathrm{mm}$, in percentage points. Positive values favor Ours. Intervals use 20,000 paired demonstration resamples (pointwise 95\% percentile intervals, no multiplicity adjustment). Scores are failure-adjusted with $N_{\rm att}=10$. Every interval excludes zero except Sharpa Wave versus \texttt{Position} and Contact PyRoki. On the matched sequences $\mathcal I$, every Allegro and Dex3 interval still excludes zero; Sharpa has no failures, so its matched intervals are identical. These intervals describe variation across the selected demonstrations, not independent policy training seeds.}
\label{tab:retarget-paired-intervals}
\begingroup
\setlength{\tabcolsep}{2pt}
\renewcommand{\arraystretch}{1.13}
\begin{tabular*}{\columnwidth}{@{\extracolsep{\fill}}lrr@{}}
\toprule
Comparison & $\Delta$F1 & 95\% interval \\
\midrule
\multicolumn{3}{@{}l}{\textit{Allegro}} \\
Ours $-$ OmniRetarget & $17.4$ & $[7.7,\,26.6]$ \\
Ours $-$ DexPilot & $17.2$ & $[11.6,\,23.3]$ \\
Ours $-$ \texttt{Position} & $8.3$ & $[1.2,\,14.7]$ \\
Ours $-$ AnyTeleop & $17.1$ & $[11.4,\,23.4]$ \\
Ours $-$ Contact PyRoki & $14.9$ & $[4.8,\,24.4]$ \\
\midrule
\multicolumn{3}{@{}l}{\textit{Unitree Dex3}} \\
Ours $-$ OmniRetarget & $30.4$ & $[18.3,\,39.8]$ \\
Ours $-$ DexPilot & $35.5$ & $[30.4,\,39.6]$ \\
Ours $-$ \texttt{Position} & $27.8$ & $[23.3,\,32.3]$ \\
Ours $-$ AnyTeleop & $35.3$ & $[30.1,\,39.6]$ \\
Ours $-$ Contact PyRoki & $28.4$ & $[14.3,\,37.4]$ \\
\midrule
\multicolumn{3}{@{}l}{\textit{Sharpa Wave}} \\
Ours $-$ OmniRetarget & $22.7$ & $[2.2,\,40.1]$ \\
Ours $-$ DexPilot & $15.3$ & $[5.5,\,23.7]$ \\
Ours $-$ \texttt{Position} & $13.3$ & $[-4.5,\,28.3]$ \\
Ours $-$ AnyTeleop & $16.2$ & $[6.7,\,24.2]$ \\
Ours $-$ Contact PyRoki & $10.5$ & $[-7.1,\,26.6]$ \\
\bottomrule
\end{tabular*}
\endgroup
\end{table}

\section{Dynamic retargeting}
\label{app:dynamic}
\subsection{Contact metrics of the dynamically retargeted grasps}
\label{app:rl-contact}
Table~\ref{tab:retargeting_comparison} scores the \emph{achieved} final state of every RL rollout, not the tracking target, with the contact protocol of Appendix~\ref{app:retarget-metrics}: location-aware, area-weighted mapped-finger contact at a $5\,\mathrm{mm}$ robot threshold and the unsigned distance from the human contact patch to the corresponding robot part, including missed contacts. The human reference is the demonstration's last retargeted frame (MANO hand within $4.5\,\mathrm{mm}$ of an object vertex, each vertex labelled with the touching finger), which is the grasp the policies are trained to reach. Each rollout replaces one trajectory frame: object surface area is pooled over the rollouts of a category, then categories are averaged with equal weight. The simulator rescales every object per rollout and per axis; the human patch is carried onto the scaled mesh by vertex index and all distances use the scaled mesh. Robot collision geometry, finger correspondence and the $2\,\mathrm{mm}$ surface sampling are those of the kinematic evaluation. Every 8th rollout of each 2048-rollout file is scored (256 per category); success rates use all rollouts.

Retargeting failures and aggregation follow Appendix~\ref{app:retarget-aggregation}: F1, precision, recall and success rate are macro-averaged over all $N_{\rm att}=10$ attempts per hand with the three confirmed OmniRetarget failures scored as zero, and geometric quantities average the $n_{\mathcal I}$ categories every method completed. Table~\ref{tab:rl-contact} gives contact F1, precision and recall on all rollouts; SR and patch distance on all rollouts are in Tables~\ref{tab:retargeting_comparison} and~\ref{tab:ablation}. Because a failed rollout usually leaves the object on the floor, far from the hand, the table also repeats F1 and patch distance on successful rollouts only, where $n_{\rm cat}$ counts the categories with at least ten successful rollouts (retargeting failures still count as zero for F1).

\textbf{Collision surface.} The simulator resolves contact against a convex decomposition of each object, which lies up to a few millimetres outside the visual mesh. As a robustness check we re-scored every cell whose decomposition was available (7 categories: alarmclock, apple, bowl, cubelarge, cup, hammer, lightbulb) with each visual vertex mapped to the closest point on the outer surface of the convex pieces and the robot measured against that point. Over the 107 cells with at least ten successful rollouts, F1 on successful rollouts changes by -3.1 to +4.6 points and patch distance by -0.6 to +0.2\,mm. Patch-distance rankings of the methods are identical on both surfaces, and F1 rankings change only between methods less than 0.4 points apart (all-rollout scores, same categories). The tables therefore use the visual mesh, which is also the surface the human patch is defined on.

\begin{table*}[t]
\centering
\footnotesize
\caption{Contact metrics of the achieved final grasps. SR and patch distance on all rollouts are in Tables~\ref{tab:retargeting_comparison} and~\ref{tab:ablation}. For successful rollouts only, $n_{\rm cat}$ counts the categories with at least ten successful rollouts behind F1 / behind patch distance. Bold and underlining mark the best and second-best distinct means within each hand and metric.}
\label{tab:rl-contact}
\begingroup
\setlength{\tabcolsep}{2.3pt}
\renewcommand{\arraystretch}{1.18}
\begin{tabular*}{\textwidth}{@{\extracolsep{\fill}}lcccccc@{}}
\toprule
 & \multicolumn{3}{c}{All rollouts} & \multicolumn{3}{c}{Successful rollouts} \\
\cmidrule(lr){2-4} \cmidrule(lr){5-7}
Method & F1 $\uparrow$ & Precision $\uparrow$ & Recall $\uparrow$ & $n_{\rm cat}$ & F1 $\uparrow$ & Patch $\downarrow$ \\
 & (\%) & (\%) & (\%) &  & (\%) & (mm) \\
\midrule
\multicolumn{7}{@{}l}{\textit{Dex3} \hfill $N_{\rm att}=10,\quad n_{\mathcal I}=8$} \\
\addlinespace[2pt]
OmniRetarget~\cite{Yang2026OmniRetarget} & \RetargetStat{5.5}{6.2} & \RetargetStat{6.8}{6.9} & \RetargetStat{5.2}{5.9} & 9/7 & \RetargetStat{7.3}{7.3} & \RetargetStat{25.3}{7.1} \\
DexPilot~\cite{handa2020dexpilot} & \RetargetStat{5.5}{8.1} & \RetargetStat{8.4}{11.3} & \RetargetStat{4.9}{7.5} & 2/2 & \RetargetStat{6.5}{8.7} & \RetargetStat{28.6}{15.4} \\
\texttt{Position}~\cite{qin2023anyteleop} & \RetargetStat{\underline{11.3}}{7.2} & \RetargetStat{\underline{15.2}}{10.4} & \RetargetStat{\underline{10.6}}{8.1} & 5/4 & \RetargetStat{11.9}{6.0} & \RetargetStat{21.1}{10.4} \\
AnyTeleop~\cite{qin2023anyteleop} & \RetargetStat{5.7}{6.5} & \RetargetStat{8.9}{9.9} & \RetargetStat{5.2}{7.0} & 3/3 & \RetargetStat{12.5}{11.1} & \RetargetStat{\underline{20.9}}{9.5} \\
Contact PyRoki~\cite{kim2025pyroki} & \RetargetStat{3.7}{5.8} & \RetargetStat{7.1}{7.3} & \RetargetStat{3.3}{5.3} & 4/4 & \RetargetStat{10.1}{6.8} & \RetargetStat{36.4}{19.7} \\
\algabbr{} w/o contact & \RetargetStat{9.0}{8.5} & \RetargetStat{13.6}{14.8} & \RetargetStat{8.7}{9.3} & 6/5 & \RetargetStat{\underline{15.7}}{5.4} & \RetargetStat{\mathbf{20.0}}{9.1} \\
\algabbr{} w/o object pose & \RetargetStat{8.6}{7.5} & \RetargetStat{12.6}{9.7} & \RetargetStat{7.0}{6.2} & 7/6 & \RetargetStat{11.5}{8.6} & \RetargetStat{28.1}{10.8} \\
\addlinespace[2pt]
\algabbr{} (Ours) & \RetargetStat{\mathbf{15.0}}{8.5} & \RetargetStat{\mathbf{18.5}}{11.1} & \RetargetStat{\mathbf{13.6}}{7.3} & 10/8 & \RetargetStat{\mathbf{16.0}}{8.9} & \RetargetStat{21.9}{6.6} \\
\midrule
\multicolumn{7}{@{}l}{\textit{Allegro} \hfill $N_{\rm att}=10,\quad n_{\mathcal I}=9$} \\
\addlinespace[2pt]
OmniRetarget~\cite{Yang2026OmniRetarget} & \RetargetStat{4.3}{3.7} & \RetargetStat{6.0}{5.8} & \RetargetStat{3.8}{3.0} & 9/8 & \RetargetStat{4.4}{3.3} & \RetargetStat{24.4}{6.5} \\
DexPilot~\cite{handa2020dexpilot} & \RetargetStat{5.7}{5.6} & \RetargetStat{\mathbf{11.5}}{9.9} & \RetargetStat{4.5}{4.0} & 6/6 & \RetargetStat{8.7}{5.2} & \RetargetStat{20.5}{7.2} \\
\texttt{Position}~\cite{qin2023anyteleop} & \RetargetStat{\underline{8.3}}{6.3} & \RetargetStat{11.2}{7.5} & \RetargetStat{\underline{7.6}}{6.7} & 9/8 & \RetargetStat{8.3}{6.0} & \RetargetStat{26.4}{16.2} \\
AnyTeleop~\cite{qin2023anyteleop} & \RetargetStat{6.5}{5.5} & \RetargetStat{\underline{11.4}}{11.1} & \RetargetStat{5.2}{4.5} & 5/5 & \RetargetStat{8.8}{4.6} & \RetargetStat{22.8}{10.4} \\
Contact PyRoki~\cite{kim2025pyroki} & \RetargetStat{2.4}{3.7} & \RetargetStat{3.6}{5.2} & \RetargetStat{2.2}{3.3} & 4/4 & \RetargetStat{7.5}{4.2} & \RetargetStat{23.8}{7.8} \\
\algabbr{} w/o contact & \RetargetStat{6.2}{6.2} & \RetargetStat{10.0}{7.7} & \RetargetStat{5.4}{5.8} & 7/6 & \RetargetStat{\underline{10.7}}{5.4} & \RetargetStat{\underline{19.9}}{5.8} \\
\algabbr{} w/o object pose & \RetargetStat{\mathbf{8.7}}{5.5} & \RetargetStat{10.3}{6.0} & \RetargetStat{\mathbf{8.6}}{6.1} & 7/6 & \RetargetStat{\mathbf{11.3}}{6.2} & \RetargetStat{\mathbf{16.7}}{3.4} \\
\addlinespace[2pt]
\algabbr{} (Ours) & \RetargetStat{6.7}{5.8} & \RetargetStat{8.0}{6.9} & \RetargetStat{6.5}{5.6} & 10/9 & \RetargetStat{8.0}{7.5} & \RetargetStat{25.2}{12.2} \\
\midrule
\multicolumn{7}{@{}l}{\textit{Sharpa} \hfill $N_{\rm att}=10,\quad n_{\mathcal I}=10$} \\
\addlinespace[2pt]
OmniRetarget~\cite{Yang2026OmniRetarget} & \RetargetStat{8.8}{8.0} & \RetargetStat{10.4}{8.1} & \RetargetStat{9.7}{10.5} & 8/8 & \RetargetStat{10.1}{8.8} & \RetargetStat{27.0}{16.6} \\
DexPilot~\cite{handa2020dexpilot} & \RetargetStat{3.6}{5.4} & \RetargetStat{7.1}{10.1} & \RetargetStat{2.7}{3.7} & 4/4 & \RetargetStat{5.1}{6.7} & \RetargetStat{30.6}{16.8} \\
\texttt{Position}~\cite{qin2023anyteleop} & \RetargetStat{3.5}{5.9} & \RetargetStat{6.0}{8.3} & \RetargetStat{3.2}{5.3} & 4/4 & \RetargetStat{8.1}{9.4} & \RetargetStat{29.0}{15.0} \\
AnyTeleop~\cite{qin2023anyteleop} & \RetargetStat{5.3}{6.5} & \RetargetStat{7.9}{9.1} & \RetargetStat{4.7}{5.3} & 7/7 & \RetargetStat{8.4}{6.4} & \RetargetStat{23.3}{7.9} \\
Contact PyRoki~\cite{kim2025pyroki} & \RetargetStat{7.7}{7.6} & \RetargetStat{9.7}{8.5} & \RetargetStat{7.5}{8.1} & 8/8 & \RetargetStat{10.5}{8.4} & \RetargetStat{29.3}{18.5} \\
\algabbr{} w/o contact & \RetargetStat{\underline{10.6}}{6.1} & \RetargetStat{\underline{13.6}}{8.3} & \RetargetStat{\underline{10.3}}{6.7} & 9/9 & \RetargetStat{\underline{12.6}}{5.4} & \RetargetStat{\underline{22.2}}{9.9} \\
\algabbr{} w/o object pose & \RetargetStat{7.1}{6.7} & \RetargetStat{7.4}{7.0} & \RetargetStat{8.0}{7.9} & 8/8 & \RetargetStat{7.6}{7.4} & \RetargetStat{28.6}{16.5} \\
\addlinespace[2pt]
\algabbr{} (Ours) & \RetargetStat{\mathbf{15.1}}{8.2} & \RetargetStat{\mathbf{17.3}}{10.5} & \RetargetStat{\mathbf{16.2}}{8.7} & 10/10 & \RetargetStat{\mathbf{15.3}}{8.4} & \RetargetStat{\mathbf{17.3}}{3.8} \\
\bottomrule
\end{tabular*}
\endgroup
\end{table*}

\subsection{Paired category-level differences}
\label{app:rl-paired}
Table~\ref{tab:rl-paired-intervals} reports paired differences between Ours
and each baseline of Table~\ref{tab:retargeting_comparison}, using object
categories as the paired unit, since every kinematic reference is refined
and evaluated on the same ten categories. For category $c$ we take
$\Delta\mathrm{SR}_c=\mathrm{SR}_{\mathrm{Ours},c}-\mathrm{SR}_{\mathrm{base},c}$
and
$\Delta\mathrm{C\text{-}ADD}_c=\mathrm{C\text{-}ADD}_{\mathrm{base},c}-\mathrm{C\text{-}ADD}_{\mathrm{Ours},c}$,
so positive values favor Ours for both metrics. We resample categories with
replacement, keeping each baseline's score paired with Ours on the same
category, with 20,000 bootstrap draws and seed 20260917, the protocol of
Table~\ref{tab:retarget-paired-intervals}. SR uses all ten categories, with
the three OmniRetarget retargeting failures scored as zero as in
Table~\ref{tab:retargeting_comparison}. C-ADD is undefined when no kinematic
reference exists, so the OmniRetarget C-ADD comparisons use only the
categories it completed ($8$ on Dex3, $9$ on Allegro, and $10$ on Sharpa).
Their mean differences therefore differ slightly from the difference of the
means in Table~\ref{tab:retargeting_comparison}, where Ours is averaged over
all ten categories.

At the nominal 5\% level, the intervals exclude zero in 11 of the 15 SR
comparisons and 10 of the 15 C-ADD comparisons. On Dex3 and Sharpa, every SR
interval excludes zero. On Allegro, only the interval against Contact PyRoki
does: Ours has its largest category-to-category spread on this hand (SD
$30.0$ points, versus $9.5$ and $3.6$ on Dex3 and Sharpa), and
\texttt{Position} and OmniRetarget come within $1.6$ and $8.6$ points of Ours
in mean SR. For C-ADD, the intervals against OmniRetarget include zero on all
three hands, as do those against \texttt{Position} on Allegro and Contact
PyRoki on Sharpa. Each category's SR is estimated from 2048 rollouts
(binomial standard error at most $1.1$ points), so the intervals are
dominated by variation between categories rather than rollout sampling. As
in Appendix~\ref{app:kinematic}, the intervals are pointwise 95\% percentile
intervals without a multiplicity adjustment across the 15 comparisons per
metric, so they do not support a joint claim that Ours outperforms every
baseline. They characterize variation across the ten selected object
categories, not variation across independent RL training seeds.

\begin{table}[t]
\centering
\footnotesize
\caption{Paired differences in residual RL dynamic retargeting between Ours
and each baseline across object categories. Positive values favor Ours:
$\Delta$SR is Ours minus baseline, and $\Delta$C-ADD is baseline minus Ours.
Intervals use 20,000 paired category resamples (pointwise 95\% percentile
intervals, no multiplicity adjustment). SR uses all ten categories with
retargeting failures scored as zero; C-ADD against OmniRetarget uses the
$8$, $9$, and $10$ categories it completed on Dex3, Allegro, and Sharpa.
These intervals describe variation across object categories, not
independent RL training seeds.}
\label{tab:rl-paired-intervals}
\begingroup
\setlength{\tabcolsep}{2pt}
\renewcommand{\arraystretch}{1.13}
\begin{tabular*}{\columnwidth}{@{\extracolsep{\fill}}lrrrr@{}}
\toprule
 & \multicolumn{2}{c}{SR (pp)} & \multicolumn{2}{c}{C-ADD ($10^{-2}$\,m)} \\
\cmidrule(lr){2-3} \cmidrule(l){4-5}
Baseline & $\Delta$ & 95\% interval & $\Delta$ & 95\% interval \\
\midrule
\multicolumn{5}{@{}l}{\textit{Dex3}} \\
OmniRetarget & $29.3$ & $[8.7,\,51.5]$ & $1.48$ & $[-0.68,\,4.91]$ \\
DexPilot & $65.2$ & $[39.9,\,85.6]$ & $6.58$ & $[2.78,\,11.30]$ \\
\texttt{Position} & $42.7$ & $[15.2,\,69.5]$ & $3.22$ & $[0.85,\,6.07]$ \\
AnyTeleop & $63.4$ & $[42.6,\,81.5]$ & $7.14$ & $[3.31,\,11.51]$ \\
Contact PyRoki & $49.2$ & $[22.8,\,73.9]$ & $8.32$ & $[2.80,\,14.82]$ \\
\midrule
\multicolumn{5}{@{}l}{\textit{Allegro}} \\
OmniRetarget & $8.6$ & $[-20.3,\,35.6]$ & $1.94$ & $[-0.41,\,4.64]$ \\
DexPilot & $30.8$ & $[-3.7,\,61.8]$ & $4.11$ & $[1.17,\,7.22]$ \\
\texttt{Position} & $1.6$ & $[-27.1,\,29.0]$ & $1.24$ & $[-1.45,\,4.58]$ \\
AnyTeleop & $32.7$ & $[-1.6,\,63.0]$ & $3.04$ & $[0.48,\,6.00]$ \\
Contact PyRoki & $44.3$ & $[9.0,\,72.9]$ & $10.62$ & $[3.90,\,18.86]$ \\
\midrule
\multicolumn{5}{@{}l}{\textit{Sharpa}} \\
OmniRetarget & $35.3$ & $[11.2,\,60.2]$ & $3.19$ & $[-0.30,\,7.19]$ \\
DexPilot & $64.5$ & $[38.5,\,86.4]$ & $7.28$ & $[2.76,\,12.18]$ \\
\texttt{Position} & $58.0$ & $[28.0,\,83.9]$ & $5.91$ & $[1.13,\,12.23]$ \\
AnyTeleop & $55.1$ & $[31.2,\,76.2]$ & $4.70$ & $[1.52,\,8.11]$ \\
Contact PyRoki & $35.6$ & $[13.9,\,58.8]$ & $2.35$ & $[-0.11,\,5.29]$ \\
\bottomrule
\end{tabular*}
\endgroup
\end{table}

\section{Real-World Experiments}
\label{app:real-world}
\begin{figure*}[t]
    \centering
    \includegraphics[width=\textwidth]{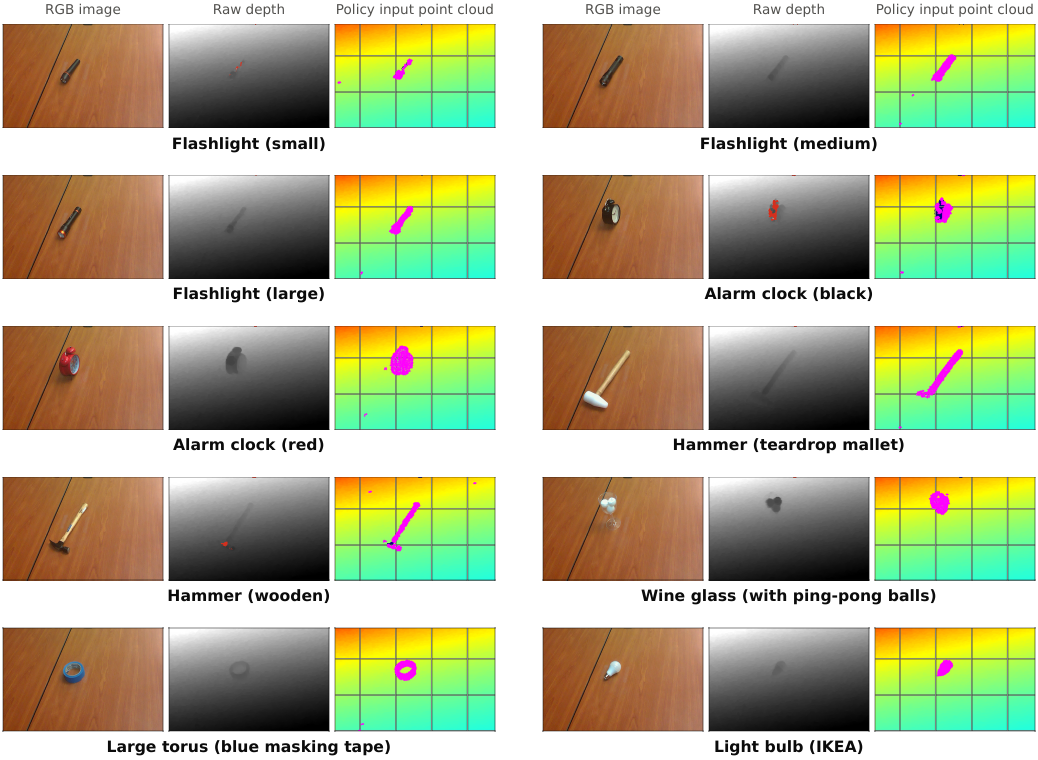}
    \caption{\textbf{Real-World Policy Inputs.} For ten of the 30 physical
    instances, the RGB image from the L515 camera (for reference only; the
    policy does not observe RGB), the raw depth image (darker is closer; red
    marks pixels with no LiDAR return), and the policy input: the 512-point
    cloud after table removal (magenta), drawn over the color-coded depth
    image.}
    \label{fig:real_world_policy_input}
\end{figure*}

Hardware evaluation uses three distinct physical objects per category, each tested at 10 poses. Nine poses are fixed at the center, four corners, and four edge midpoints of the
$10\,\mathrm{cm}\times10\,\mathrm{cm}$ pose-randomization region. The tenth pose provides an additional stress test: for
non-rotationally symmetric objects, we sample a random position with an orientation offset of at least $15^\circ$; for rotationally symmetric objects, we instead place the object slightly outside the pose-randomization region. This yields 30 trials per category and 300 trials overall. The hardware macro mean is 89.33\% (268/300).
Tables~\ref{tab:real_objects_1} and~\ref{tab:real_objects_2} list every
physical instance with its photo, mass, and dimensions, alongside the
dimensions of the GRAB object on which the category's policy was trained and
the per-instance success rate.

\providecommand{\objphoto}[1]{%
    \raisebox{-0.5\height}{\includegraphics[height=1.2cm]{#1}}%
}
\providecommand{\objrowshift}{-0.9cm}

\begin{table*}[!t]
\centering
\caption{\textbf{Real-World Object Instances (I).}
The three physical instances of the first five object categories used in the hardware evaluation.
For each instance, we report its photo, hand-measured mass and dimensions
(length $\times$ width $\times$ height), and pick-up success rate over the 10 test poses.
For each category, we additionally report the dimensions of the GRAB object mesh used to
train the policy in simulation (before anisotropic scale randomization) and the success
rate over all 30 trials. Simulated object masses are sampled in $[50,400]$\,g.}
\label{tab:real_objects_1}

\footnotesize
\setlength{\tabcolsep}{5pt}

\begin{tabular*}{\textwidth}{
    @{\extracolsep{\fill}}
    llcrrrrcc c
    @{}
}
\toprule
\multirow{2}{*}{Category}
& \multirow{2}{*}{Instance}
& \multirow{2}{*}{Photo}
& \multicolumn{4}{c}{Physical instance}
& \multirow{2}{*}{\shortstack{Reference object\\dimensions (cm)}}
& \multicolumn{2}{c}{Success rate} \\
\cmidrule(lr){4-7}
\cmidrule(l){9-10}
& & &
Mass (g) & L (cm) & W (cm) & H (cm)
& & Instance & Category \\
\midrule

\multirow{3}{*}[\objrowshift]{Alarm clock}
& Red
& \objphoto{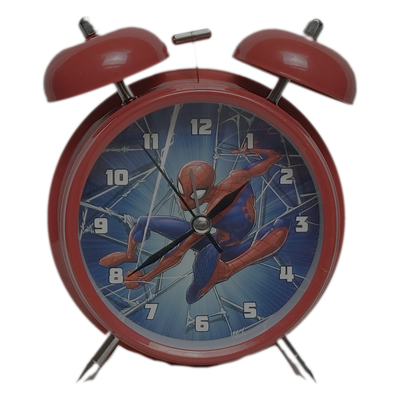}
& 225 & 11.7 & 5.6 & 13.4
& \multirow{3}{*}[\objrowshift]{$12.6 \times 4.5 \times 10.8$}
& 7/10
& \multirow{3}{*}[\objrowshift]{26/30} \\[3pt]

& Black
& \objphoto{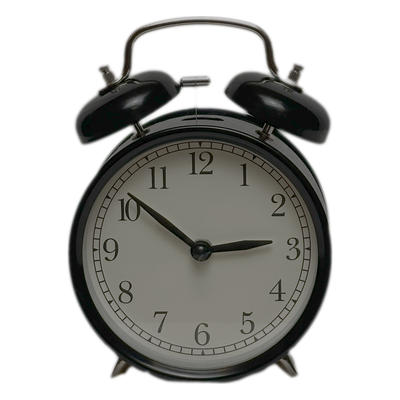}
& 240 & 9.7 & 5.3 & 13.6
& & 9/10 & \\[3pt]

& Swivel
& \objphoto{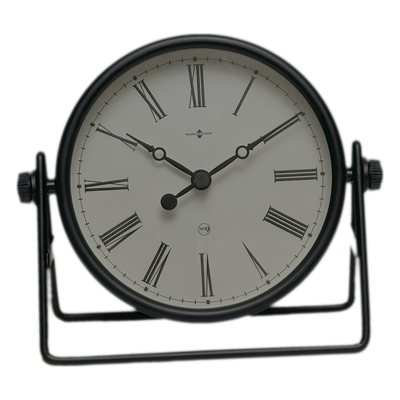}
& 257 & 12.1 & 4.3 & 12.7
& & 10/10 & \\[3pt]

\midrule

\multirow{3}{*}[\objrowshift]{Cup}
& Short beige
& \objphoto{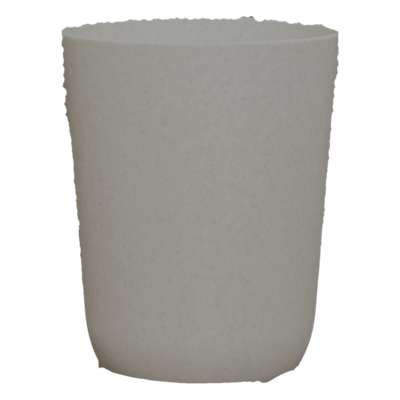}
& 47 & 7.9 & 7.9 & 10.6
& \multirow{3}{*}[\objrowshift]{$8.5 \times 8.5 \times 10.0$}
& 10/10
& \multirow{3}{*}[\objrowshift]{28/30} \\[3pt]

& Tall beige
& \objphoto{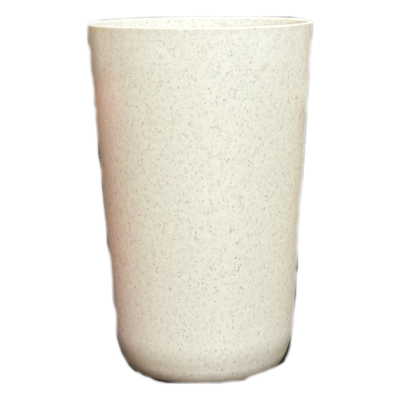}
& 62 & 7.8 & 7.8 & 14.5
& & 9/10 & \\[3pt]

& Soda can
& \objphoto{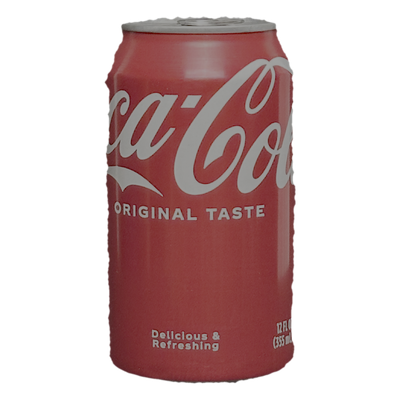}
& 382 & 6.6 & 6.6 & 12.2
& & 9/10 & \\[3pt]

\midrule

\multirow{3}{*}[\objrowshift]{Cube}
& Kleenex box
& \objphoto{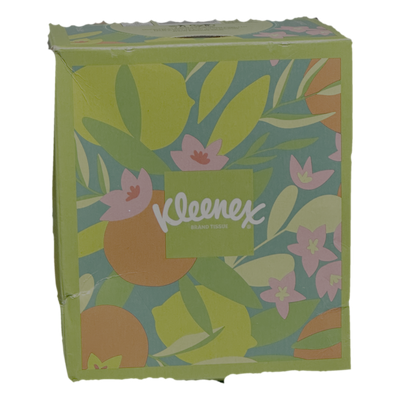}
& 141 & 11.2 & 11.0 & 12.6
& \multirow{3}{*}[\objrowshift]{$12.0 \times 12.0 \times 12.0$}
& 9/10
& \multirow{3}{*}[\objrowshift]{26/30} \\[3pt]

& Quest box
& \objphoto{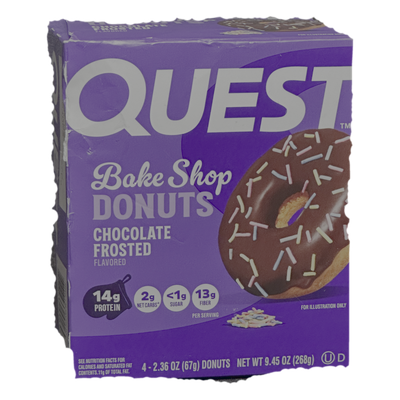}
& 316 & 10.1 & 10.9 & 12.6
& & 9/10 & \\[3pt]

& Febreze box
& \objphoto{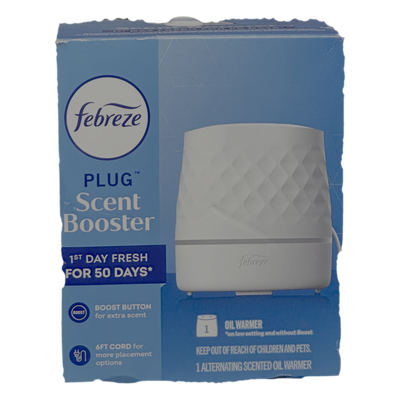}
& 257 & 10.5 & 11.2 & 13.6
& & 8/10 & \\[3pt]

\midrule

\multirow{3}{*}[\objrowshift]{Bowl}
& Grey
& \objphoto{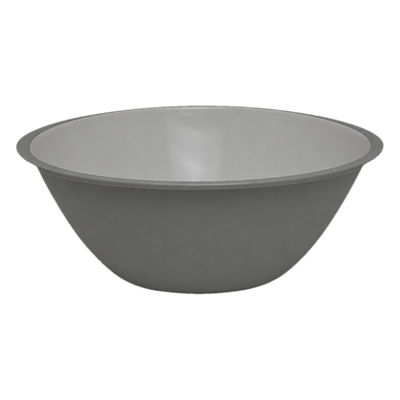}
& 152 & 17.2 & 17.2 & 6.6
& \multirow{3}{*}[\objrowshift]{$14.9 \times 14.9 \times 7.7$}
& 10/10
& \multirow{3}{*}[\objrowshift]{27/30} \\[3pt]

& Large wooden
& \objphoto{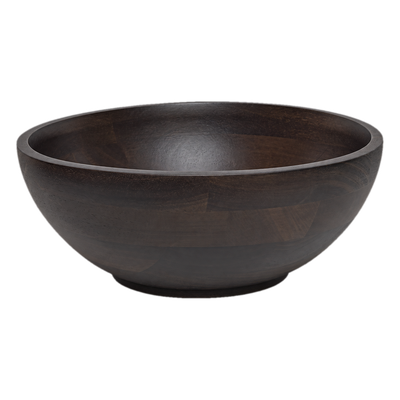}
& 264 & 20.2 & 20.2 & 7.8
& & 8/10 & \\[3pt]

& Green
& \objphoto{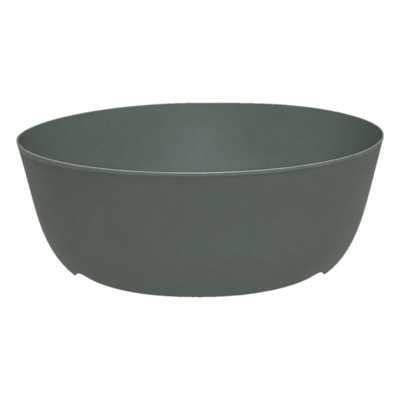}
& 78 & 16.5 & 16.5 & 6.7
& & 9/10 & \\[3pt]

\midrule

\multirow{3}{*}[\objrowshift]{Apple}
& Apple 1
& \objphoto{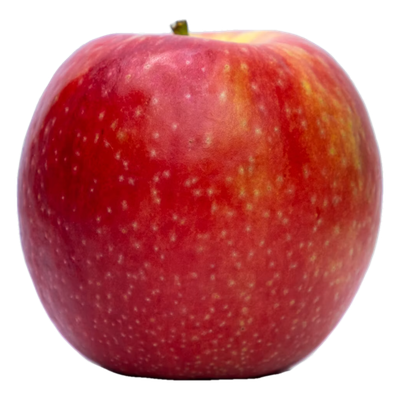}
& 223 & 8.1 & 8.2 & 8.2
& \multirow{3}{*}[\objrowshift]{$8.7 \times 8.2 \times 9.8$}
& 9/10
& \multirow{3}{*}[\objrowshift]{26/30} \\[3pt]

& Apple 2
& \objphoto{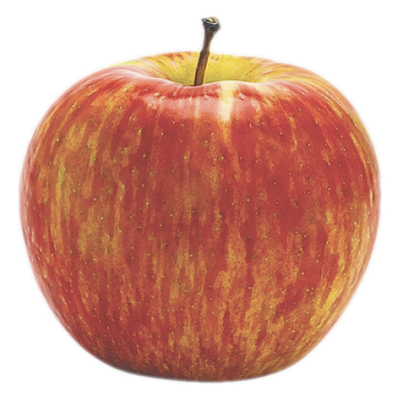}
& 265 & 8.7 & 8.4 & 8.5
& & 8/10 & \\[3pt]

& Apple 3
& \objphoto{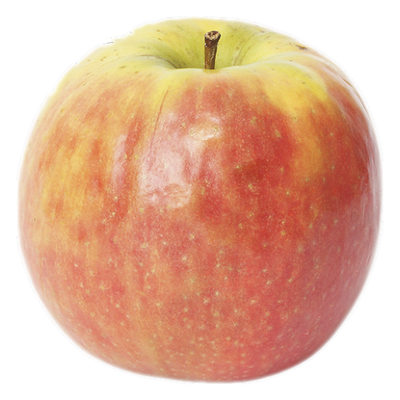}
& 247 & 8.5 & 8.4 & 8.3
& & 9/10 & \\[3pt]

\bottomrule
\end{tabular*}
\end{table*}

\begin{table*}[!t]
\centering
\caption{\textbf{Real-World Object Instances (II).}
The three physical instances of the remaining five object categories; columns are as in
Table~\ref{tab:real_objects_1}. The hammer, flashlight, and light bulb instances lie flat
on the table; the light bulb dimensions are reported with its long axis as the height.}
\label{tab:real_objects_2}

\footnotesize
\setlength{\tabcolsep}{5pt}

\begin{tabular*}{\textwidth}{
    @{\extracolsep{\fill}}
    llcrrrrcc c
    @{}
}
\toprule
\multirow{2}{*}{Category}
& \multirow{2}{*}{Instance}
& \multirow{2}{*}{Photo}
& \multicolumn{4}{c}{Physical instance}
& \multirow{2}{*}{\shortstack{Reference object\\dimensions (cm)}}
& \multicolumn{2}{c}{Success rate} \\
\cmidrule(lr){4-7}
\cmidrule(l){9-10}
& & &
Mass (g) & L (cm) & W (cm) & H (cm)
& & Instance & Category \\
\midrule

\multirow{3}{*}[\objrowshift]{Light bulb}
& IKEA
& \objphoto{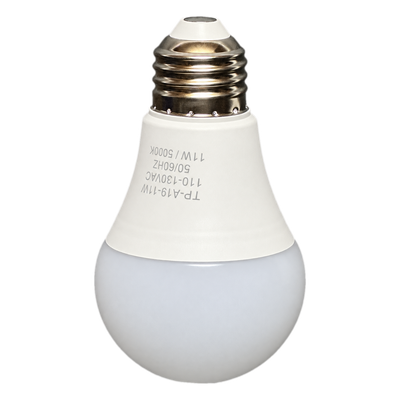}
& 26 & 6.0 & 6.0 & 10.5
& \multirow{3}{*}[\objrowshift]{$6.2 \times 6.2 \times 11.0$}
& 10/10
& \multirow{3}{*}[\objrowshift]{30/30} \\[3pt]

& Triangle
& \objphoto{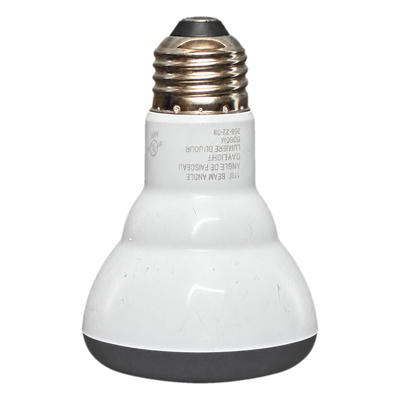}
& 42 & 6.4 & 6.4 & 9.7
& & 10/10 & \\[3pt]

& Teeth
& \objphoto{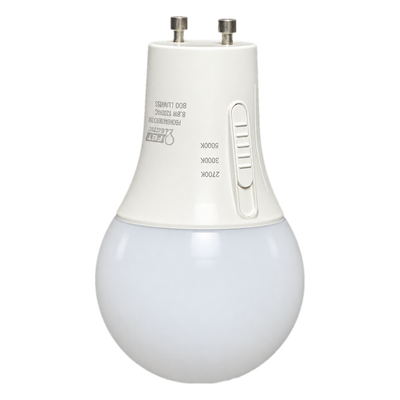}
& 41 & 6.0 & 6.0 & 10.0
& & 10/10 & \\[3pt]

\midrule

\multirow{3}{*}[\objrowshift]{Wine glass}
& Safeway
& \objphoto{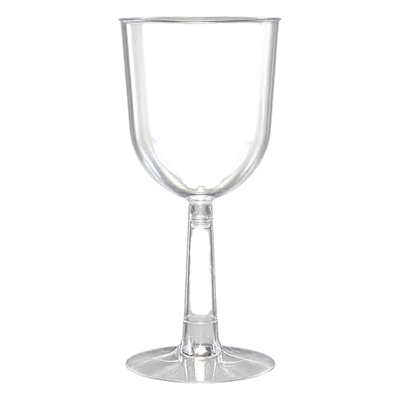}
& 31 & 4.6 & 4.6 & 17.8
& \multirow{3}{*}[\objrowshift]{$7.1 \times 7.1 \times 17.2$}
& 10/10
& \multirow{3}{*}[\objrowshift]{28/30} \\[3pt]

& Champagne flute
& \objphoto{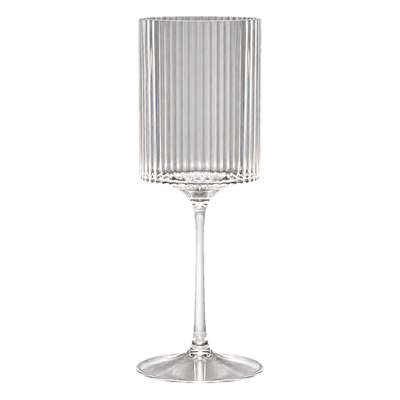}
& 47 & 3.1 & 3.1 & 19.4
& & 10/10 & \\[3pt]

& Martini
& \objphoto{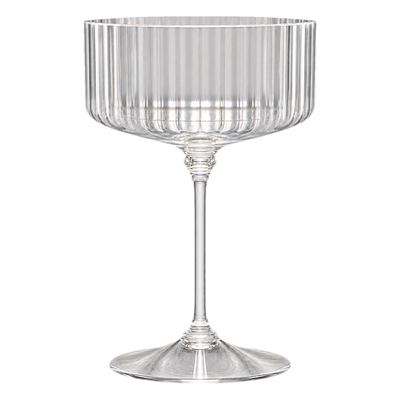}
& 42 & 6.1 & 6.1 & 11.1
& & 8/10 & \\[3pt]

\midrule

\multirow{3}{*}[\objrowshift]{Hammer}
& Red
& \objphoto{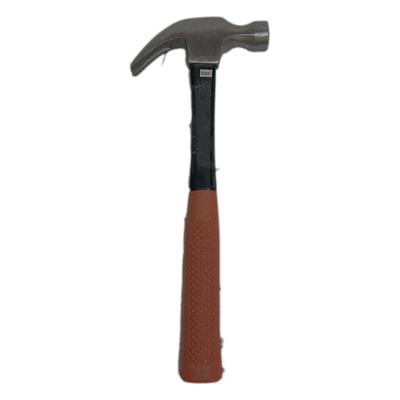}
& 452 & 28.1 & 3.1 & 2.6
& \multirow{3}{*}[\objrowshift]{$20.5 \times 11.7 \times 2.3$}
& 9/10
& \multirow{3}{*}[\objrowshift]{27/30} \\[3pt]

& Wooden
& \objphoto{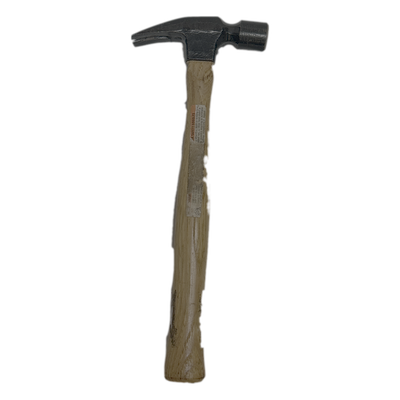}
& 380 & 28.8 & 3.2 & 2.6
& & 9/10 & \\[3pt]

& Teardrop mallet
& \objphoto{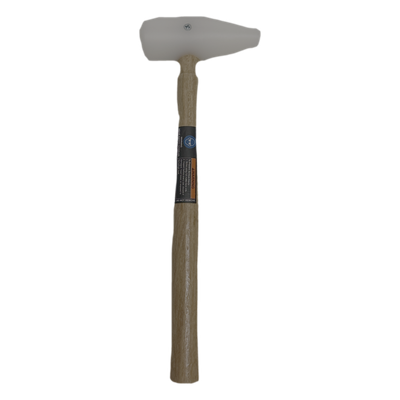}
& 260 & 33.6 & 3.2 & 3.2
& & 9/10 & \\[3pt]

\midrule

\multirow{3}{*}[\objrowshift]{Torus}
& Shoe
& \objphoto{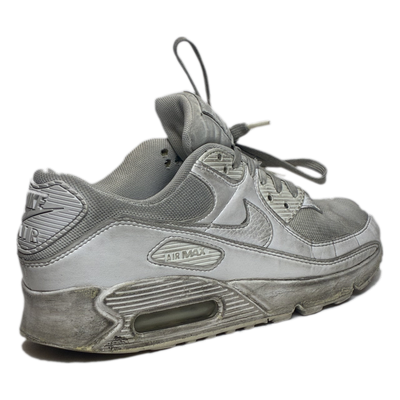}
& 330 & 26.0 & 6.7 & 9.3
& \multirow{3}{*}[\objrowshift]{$12.1 \times 12.1 \times 4.0$}
& 10/10
& \multirow{3}{*}[\objrowshift]{24/30} \\[3pt]

& Green tape (tall)
& \objphoto{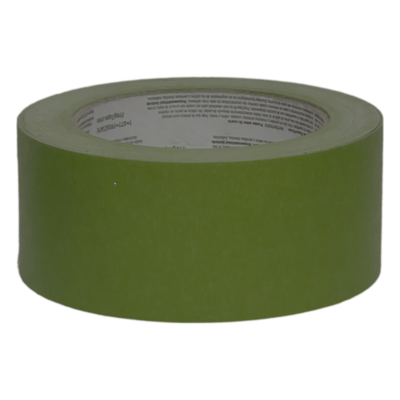}
& 230 & 11.3 & 11.3 & 4.9
& & 7/10 & \\[3pt]

& Blue tape (short)
& \objphoto{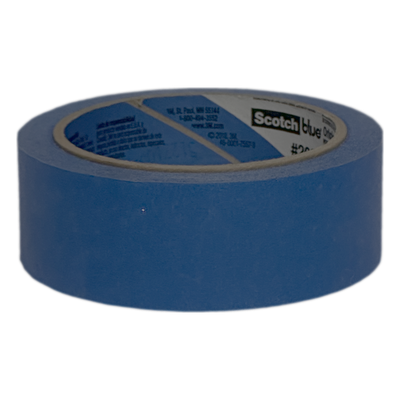}
& 120 & 10.3 & 10.3 & 3.6
& & 7/10 & \\[3pt]

\midrule

\multirow{3}{*}[\objrowshift]{Flashlight}
& Large
& \objphoto{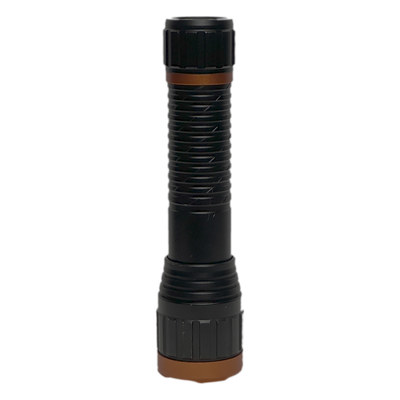}
& 253 & 3.6 & 3.6 & 17.7
& \multirow{3}{*}[\objrowshift]{$3.4 \times 3.4 \times 13.9$}
& 10/10
& \multirow{3}{*}[\objrowshift]{26/30} \\[3pt]

& Medium
& \objphoto{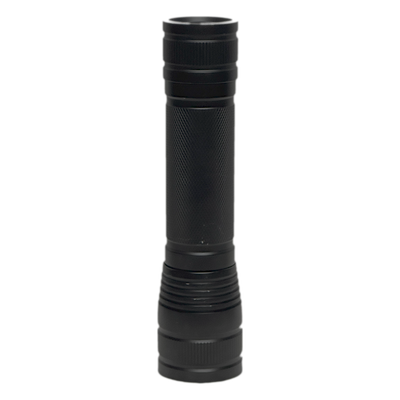}
& 270 & 3.5 & 3.5 & 17.2
& & 9/10 & \\[3pt]

& Small
& \objphoto{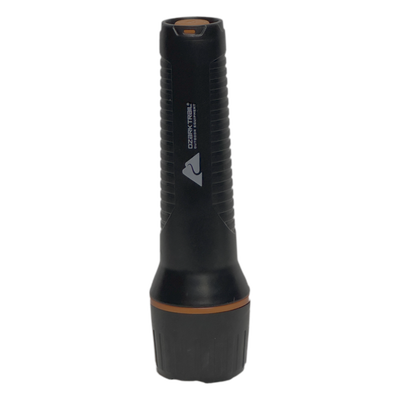}
& 89 & 2.7 & 2.7 & 14.1
& & 7/10 & \\[3pt]

\bottomrule
\end{tabular*}
\end{table*}

\textbf{Observability of Different Objects.} Figure~\ref{fig:real_world_policy_input} shows what
the visuomotor policy observes on hardware for ten of the 30 physical
instances. Depth returns are missing on parts of dark objects, such as the
small flashlight, the black alarm clock, and the head of the wooden hammer,
so their point clouds are incomplete. Only a sparse set of points remains
on the small flashlight, yet the policy still grasps it in 7 of 10 trials.
The glass of the wineglass returns no
depth at all, so the ping-pong balls placed in its bowl are the only observed
part of the object (Sec.~\ref{sec:eval-setup}). Although its stem, base, and
glass are never observed, the wineglass policy succeeds in 28 of 30 trials
across the three instances. We attribute this robustness to partial
observations to the point-cloud randomization during student training
(Sec.~\ref{sec:distillation}), which replaces points with synthetic table
residue and cable returns so that the hand and object occupy a varying and
reduced share of the 512-point cloud.

\begin{figure*}[t]
    \centering
    \includegraphics[width=\textwidth]{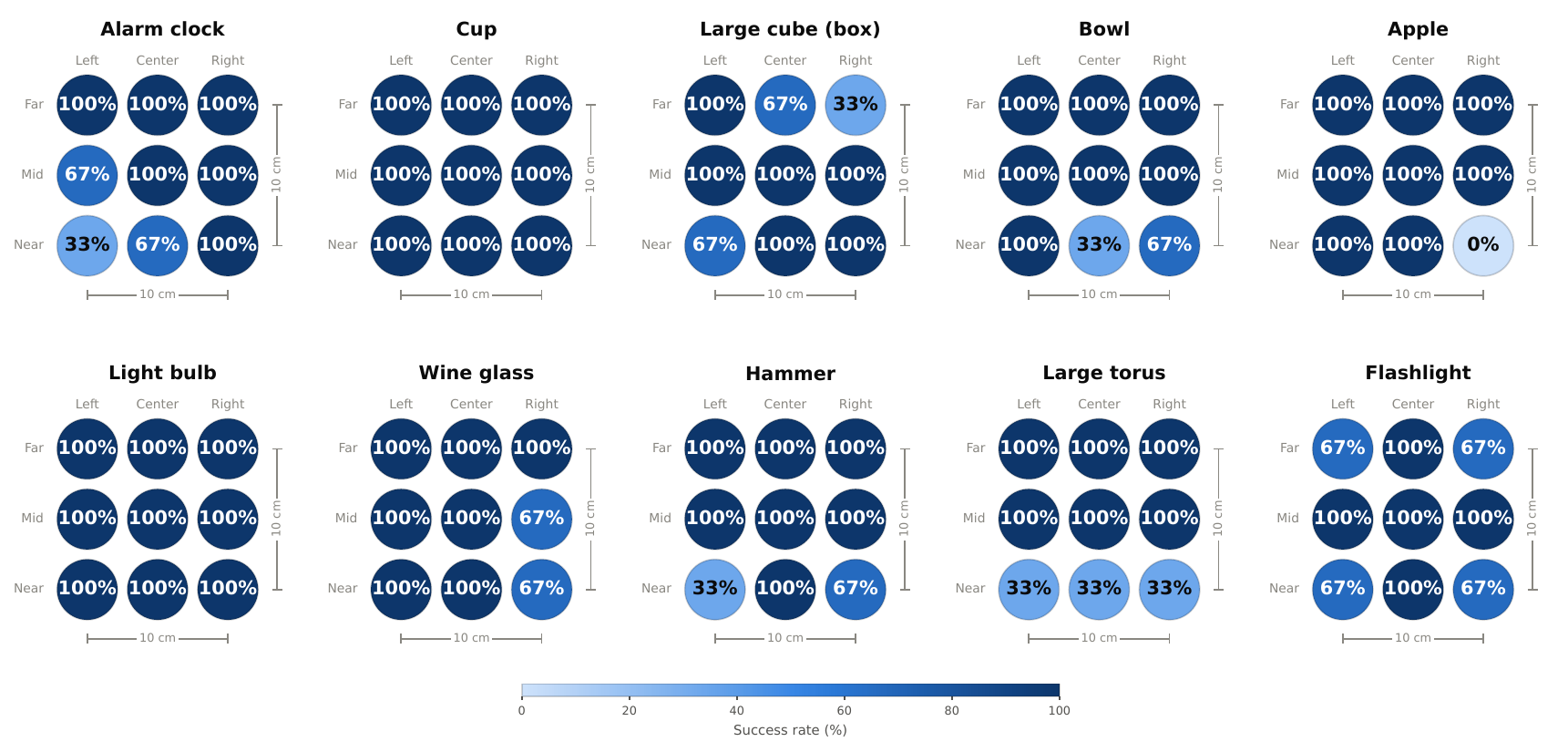}
    \caption{\textbf{Real-World Success Rate by Initial Object Location.}
    Each panel shows one object category. The nine circles are the grid
    locations spanning the $10\,\text{cm}\times10$\,cm randomization region, viewed from
    above. The near row is closest to the robot arm base, and the fingers of
    the hand point in the far direction. Each circle is colored
    and labeled by the success rate over the category's three physical
    instances at that location (3 trials per location, 27 of the 30 trials
    per category). The tenth test pose of each instance is not a grid
    location and is omitted.}
    \label{fig:real_world_sr_grid}
\end{figure*}

\textbf{Success by initial location.} Figure~\ref{fig:real_world_sr_grid}
breaks down the hardware results by the initial object location. For each
instance, nine of the ten test poses lie on a $3\times3$ grid spanning the
$10\times10$\,cm randomization region; the tenth pose lies off the grid and is
omitted from the figure. The on-grid trials succeed in 241 of 270 cases and the
off-grid trials in 27 of 30. Consistent with the
pose-dependent localization failures discussed in
Sec.~\ref{sec:distillation_experiments}, failures depend strongly on location:
22 of the 29 on-grid failures occur in the near row, 5 in the far row, and only
2 in the mid row, which contains the center of the region. In several
categories, the failures are confined to a single region. All six torus
failures occur in the near row, and all three on-grid apple failures occur at
the near-right location. The alarm clock failures cluster around the near-left
corner, the hammer failures lie in the near row, and the wineglass failures lie
in the right column. The cup and light bulb succeed at every grid location; the
cup's only two failures occur at the off-grid pose.

\end{document}